\documentclass[11pt]{article}
\usepackage{graphicx,array,amssymb,psfrag,fullpage,multirow}
\usepackage{psfrag,graphics,epsfig,multirow,multicol,times,color}
\usepackage{algorithm}
\usepackage{algorithmic}

\usepackage[agsmcite]{harvard}

\citationmode{abbr}
\definecolor{webblue}{rgb}{0,0,1}
\definecolor{webgreen}{rgb}{0,.5,0}
\definecolor{webbrown}{rgb}{.6,0,0}

\title{Predicting Abnormal Returns From News Using Text Classification}

\author{Ronny Luss\thanks{ORFE Department, Princeton University,
Princeton, NJ 08544. \texttt{rluss@princeton.edu}} \and Alexandre
d'Aspremont\thanks{ORFE Department, Princeton University, Princeton,
NJ 08544. \texttt{aspremon@princeton.edu}}}

\newcommand{\BEAS}{\begin{eqnarray*}}
\newcommand{\EEAS}{\end{eqnarray*}}
\newcommand{\BEA}{\begin{eqnarray}}
\newcommand{\EEA}{\end{eqnarray}}
\newcommand{\BEQ}{\begin{equation}}
\newcommand{\EEQ}{\end{equation}}
\newcommand{\BIT}{\begin{itemize}}
\newcommand{\EIT}{\end{itemize}}
\newcommand{\BNUM}{\begin{enumerate}}
\newcommand{\ENUM}{\end{enumerate}}

\newcommand{\BA}{\begin{array}}
\newcommand{\EA}{\end{array}}
\newcommand{\BC}{\begin{center}}
\newcommand{\EC}{\end{center}}

\newcommand{\reals}{{\mbox{\bf R}}}

\newcommand{\symm}{{\mbox{\bf S}}}  

\newcommand{\QED}{~~\rule[-1pt]{6pt}{6pt}}

\newcommand{\argmin}{\mathop{\rm argmin}}

\newtheorem{remark}[theorem]{Remark}

\newcounter{exno}

{\begin{quote}}{\end{quote}}

\makeatletter
\long\def\@makecaption#1#2{
   \vskip 9pt
   \begin{small}
   \setbox\@tempboxa\hbox{{\bf #1:} #2}
   \ifdim \wd\@tempboxa > 5.5in
        \begin{center}
        \begin{minipage}[t]{5.5in}
        \addtolength{\baselineskip}{-0.95pt}
        {\bf #1:} #2 \par
        \addtolength{\baselineskip}{0.95pt}
        \end{minipage}
        \end{center}
   \else
    \hbox to\hsize{\hfil\box\@tempboxa\hfil}
   \fi
   \end{small}\par
}
\makeatother

\newcounter{oursection}

\newcounter{lecture}

\begin{document}

\maketitle

\begin{abstract}
We show how text from news articles can be used to predict intraday price movements of financial assets using support vector machines. Multiple kernel learning is used to combine equity returns with text as predictive features to increase classification performance and we develop an analytic center cutting plane method to solve the kernel learning problem efficiently. We observe that while the direction of returns is not predictable using either text or returns, their size is, with text features producing significantly better performance than historical returns alone.
\end{abstract}

\section{Introduction}
Asset pricing models often describe the arrival of novel information by a jump process, but the characteristics of the underlying jump process are only coarsely, if at all, related to the underlying source of information. Similarly, time series models such as ARCH and GARCH have been developed to forecast volatility using asset returns data but these methods also ignore one key source of market volatility: financial news.
Our objective here is to show that text classification techniques allow a much more refined analysis of the impact of news on asset prices.

Empirical studies that examine stock return predictability can be traced back to \citeasnoun{Fama65} among others, who showed that there is no significant autocorrelation in the daily returns of thirty stocks from the Dow-Jones Industrial Average. Similar studies were conducted by \citeasnoun{Tayl1986} and \citeasnoun{Ding93}, who find significant autocorrelation in squared and absolute returns (i.e. volatility). These effects are also observed on intraday volatility patterns as demonstrated by \citeasnoun{Wood85} and by \citeasnoun{Ande97} on absolute returns.  These findings tend to demonstrate that, given solely historical stock returns, future stock returns are not predictable while volatility is. The impact of news articles has also been studied extensively. \citeasnoun{Eder93} for example studied price fluctuations in interest rate and foreign exchange futures markets following macroeconomic announcements and showed that prices mostly adjusted within one minute of major announcements. \citeasnoun{Mitc94} aggregated daily announcements by \emph{Dow Jones \& Company} into a single variable and  found no correlation with market absolute returns and weak correlation with firm-specific absolute returns.  However, \citeasnoun{Kale04} aggregated intraday news concerning companies listed on the Australian Stock Exchange into an exogenous variable in a GARCH model and found significant predictive power. These findings are attributed to the conditioning of volatility on news. Results were further improved by restricting the type of news articles included.

The most common techniques for forecasting volatility are often based on Autoregressive Conditional Heteroskedasticity (ARCH) and Generalized ARCH (GARCH) models mentioned above.  For example, intraday volatility in foreign exchange and equity markets is modeled with MA-GARCH in \citeasnoun{Ande97} and ARCH in \citeasnoun{Tayl97}.  See \citeasnoun{Boll92} for a survey of ARCH and GARCH models and various other applications.  Machine learning techniques such as neural networks and support vector machines have also been used to forecast volatility. Neural networks are used in \citeasnoun{Mall1996} to forecast implied volatility of options on the SP100 index, and support vector machines are used to forecast volatility of the SP500 index using daily returns in \citeasnoun{Gavr2006}.

Here, we show that information from press releases can be used to predict intraday abnormal returns with relatively high accuracy. Consistent with \citeasnoun{Tayl1986} and \citeasnoun{Ding93}, however, the direction of returns is not found to be predictable. We form a text classification problem where press releases are labeled positive if the absolute return jumps at some (fixed) time after the news is made public. Support vector machines (SVM) are used to solve this classification problem using both equity returns and word frequencies from press releases.  Furthermore, we use multiple kernel learning (MKL) to optimally combine equity returns with text as predictive features and increase classification performance.

Text classification is a well-studied problem in machine learning, (\citeasnoun{Duma1998} and \citeasnoun{Joac2002} among many others show that SVM significantly outperform classic methods such as naive bayes). Initially, naive bayes classifiers were used in \citeasnoun{Wuth1998} to do three-class classification of an index using daily returns for labels.  News is taken from several sources such as \emph{Reuters} and \emph{The Wall Street Journal}. Five-class classification with naive bayes classifiers is used in \citeasnoun{Lavr2000} to classify intraday price trends when articles are published at the \emph{YAHOO!Finance} website.  Support vector machines were also used to classify intraday price trends in \citeasnoun{Fung2003} using \emph{Reuters} articles and in \citeasnoun{Mitt2006b} to do four-class classification of stock returns using press releases by \emph{PRNewswire}.  Text classification has also been used to directly predict volatility (see \citeasnoun{Mitt2006a} for a survey of trading systems that use text).  Recently, \citeasnoun{Robe2007} used SVM to predict if articles from the \emph{Bloomberg service} are followed by abnormally large volatility; articles deemed important are then aggregated into a variable and used in a GARCH model similar to \citeasnoun{Kale04}. \citeasnoun{Koga2009} use Support Vector Regression (SVR) to forecast stock return volatility based on text in SEC mandated 10-K reports.  They found that reports published after the Sarbanes-Oxley Act of 2002 improved forecasts over baseline methods that did not use text.  Generating trading rules with genetic programming (GP) is another way to incorporate text for financial trading systems.  Trading rules are created in \citeasnoun{Demp2001} using GP for foreign exchange markets based on technical indicators and extended in \citeasnoun{Demp2004} to combine technical indicators with non-publicly available information.  Ensemble methods were used in \citeasnoun{Thom2003} on top of GP to create rules based on headlines posted on \emph{Yahoo} internet message boards.

Our contribution here is twofold.  First, abnormal returns are predicted using text classification techniques similar to \citeasnoun{Mitt2006b}.  Given a press release, we predict whether or not an abnormal return will occur in the next $10, 20,..., 250$ minutes using text and past absolute returns.  The algorithm in \citeasnoun{Mitt2006b} uses text to predict whether returns jump up 3\%, down 3\%, remain within these bounds, or are ``unclear'' within 15 minutes of a press release.  They consider a nine months subset of the eight years of press releases used here.  Our experiments analyze predictability of absolute returns at many horizons and demonstrate significant initial intraday predictability that decreases throughout the trading day. Second, we optimally combine text information with asset price time series to significantly enhance classification performance using multiple kernel learning (MKL). We use an analytic center cutting plane method (ACCPM) to solve the resulting MKL problem. ACCPM is particularly efficient on problems where the objective function and gradient are hard to evaluate but whose feasible set is simple enough so that analytic centers can be computed efficiently. Furthermore, because it does not suffer from conditioning issues, ACCPM can achieve higher precision targets than other first-order methods.

The rest of the paper is organized as follows.  Section \ref{sec:single_features} details the text classification problem we solve here and provides predictability results using using either text or absolute returns as features.  Section \ref{sec:multiple_features} describes the multiple kernel learning framework and details the analytic center cutting plane algorithm used to solve the resulting optimization problem. Finally, we use MKL to enhance the prediction performance.

\section{Predictions with support vector machines}
\label{sec:single_features}
Here, we describe how support vector machines can be used to make binary predictions on equity returns.  The experimental setup follows with results that use text and stock return data separately to make predictions.
\subsection{Support vector machines}
\label{sec:svm}
\begin{figure}[h!]
\BC
     \includegraphics[width=0.45 \textwidth]{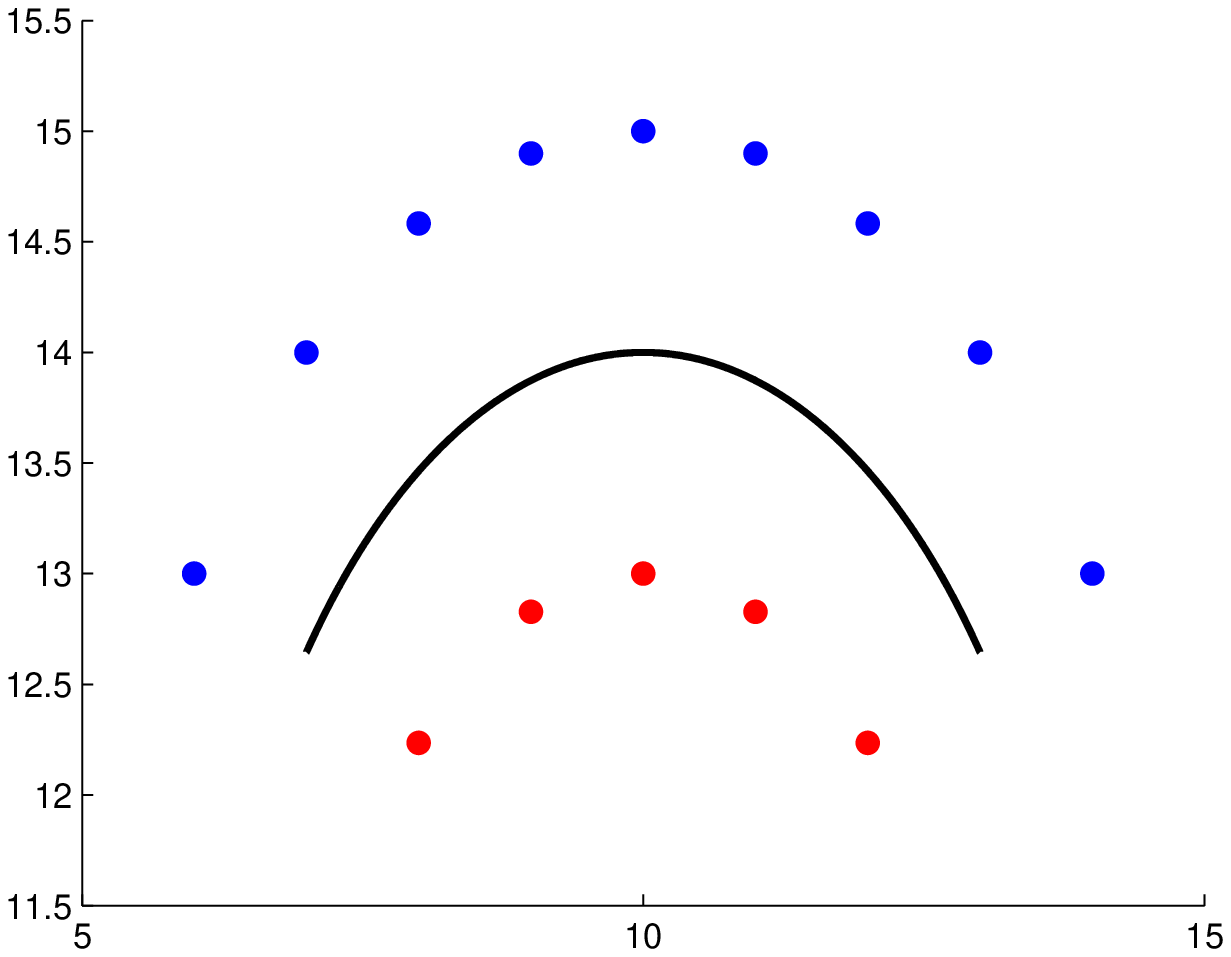}
     \includegraphics[width=0.45 \textwidth]{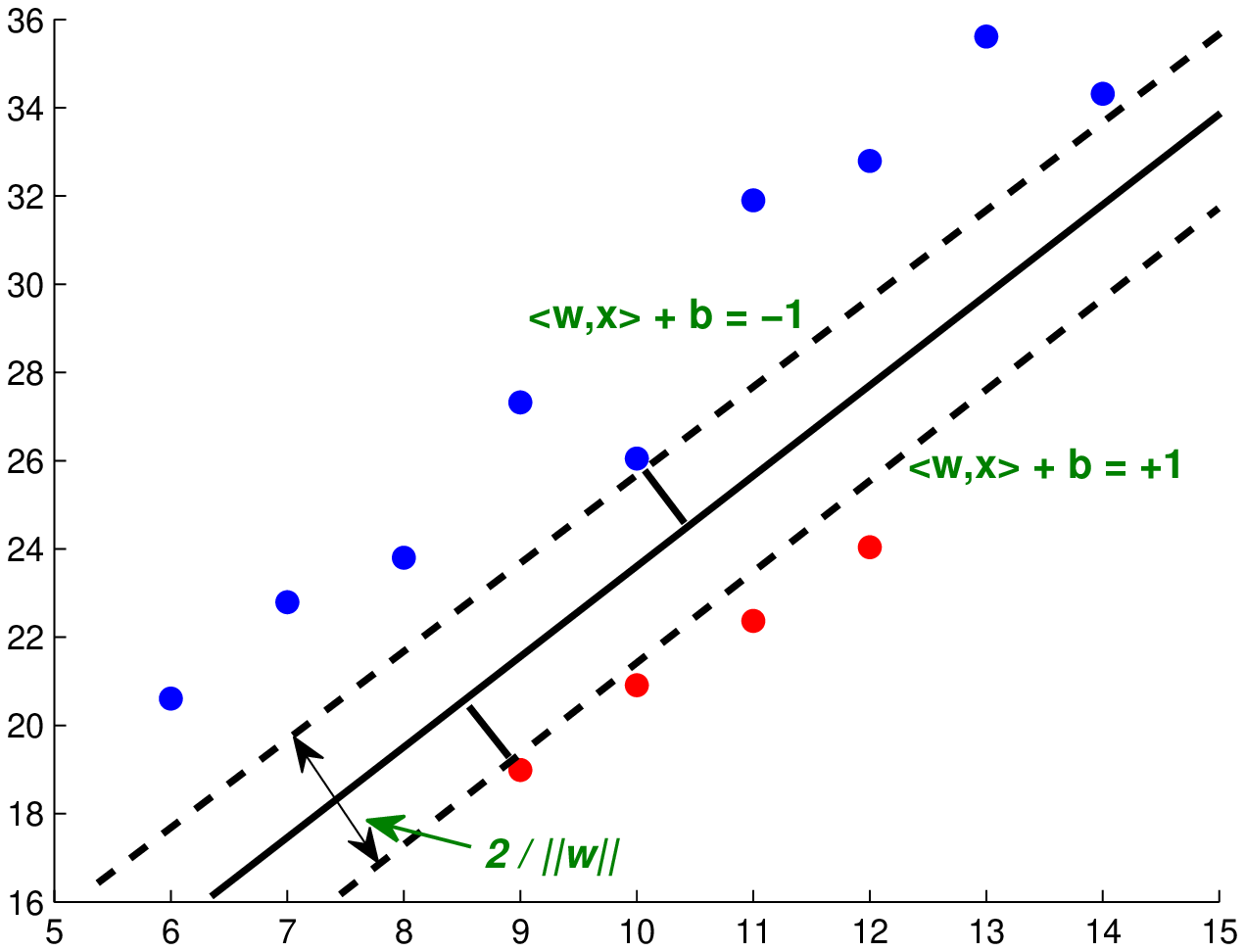}
\EC
\caption{Input Space vs. Feature Space.  For nonlinear classification, data is mapped from the input space to the feature space.  Linear classification is performed by support vector machines on mapped data in the feature space.}
\label{fig:input_feat_space}
\end{figure}

Support vector machines (SVMs) form a linear classifier by maximizing the distance, known as the \emph{margin}, between two parallel hyperplanes which separate two groups of data (see \citeasnoun{Cris2000} for a detailed reference on SVM).  This is illustrated in Figure \ref{fig:input_feat_space} (right) where the linear classifier, defined by the hyperplane $\langle w,x\rangle +b=0$, is midway between the separating hyperplanes.  Given a linear classifier, the margin can be computed explicitly as $\frac{2}{\|w\|}$ so finding the maximum margin classifier can be formulated as the linearly constrained quadratic program
\BEQ \BA{lll}
&\mbox{minimize} & \frac{1}{2}\|w\|^2+C\sum\limits_{i=1}^l{\epsilon_i}\\
&\mbox{subject to} & y_i(\langle w,\Phi(x_i)\rangle+b) \geq 1-\epsilon_i\\
& & \epsilon_i \geq 0 \EA \label{eq:svm_primal} \EEQ
in the variables $w\in\reals^d$, $b\in\reals$, and $\epsilon\in\reals^l$ where $x_i\in\reals^d$ is the $i^{th}$ data point with $d$ features, $y_i\in\{-1,1\}$ is its label, and there are $l$ points.  The first constraint dictates that points with equivalent labels are on the same side of the line.  The slack variable $\epsilon$ allows data to be misclassified while being penalized at rate $C$ in the objective, so SVMs also handle nonseparable data.  The optimal objective value in (\ref{eq:svm_primal}) can be viewed as an upper bound on the probability of misclassification for the given task.

These results can be readily extended to nonlinear classification.  Given a nonlinear classification task, the function $\Phi:x\rightarrow\Phi(x)$ maps data from an input space (Figure \ref{fig:input_feat_space} left) to a linearly separable feature space (Figure \ref{fig:input_feat_space} right) where linear classification is performed.  Problem (\ref{eq:svm_primal}) becomes numerically difficult in high dimensional feature spaces but, crucially, the complexity of solving its dual
\BEQ  \BA{lll}
&\mbox{maximize} & \alpha^Te-
\frac{1}{2}\alpha^T\mbox{diag}(y)K\mbox{diag}(y)\alpha\\
&\mbox{subject to} & \alpha^Ty=0\\
& & 0\leq\alpha\leq C \EA \label{eq:svm_dual} \EEQ
in the variables $\alpha\in\reals^l$, does not depend on the dimension of the feature space. The input to problem (\ref{eq:svm_dual}) is now an $l\times l$ matrix $K$ where $K_{ij}=\langle \Phi(x_i),\Phi(x_j)\rangle$.  Given $K$, the mapping $\Phi$ need not be specified, hence this $l$-dimensional linearly constrained quadratic program does not suffer from the high (possibly infinite) dimensionality of the mapping $\Phi$.  An explicit classifier can be constructed as function of $K$
\BEQ f(x)=\mbox{sgn}(\sum_{i=1}^l{y_i\alpha^*_iK(x_i,x)+b^*}) \label{eq:svm_classifier} \EEQ
where $x_i$ is the $i^{th}$ training sample in input space, $\alpha^*$ solves (\ref{eq:svm_dual}), and $b^*$ is computed from the KKT conditions of problem (\ref{eq:svm_primal}).

The data features are entirely described by the matrix $K$, which is called a kernel and must satisfy $K\succeq 0$, i.e. $K$ is positive-semidefinite (this is called Mercer's condition in machine learning). If $K\succeq 0$, then there exists a mapping $\Phi$ such that $K_{ij}=\langle \Phi(x_i),\Phi(x_j)\rangle$.  Thus, SVMs only require as input a \emph{kernel} function $k:(x_i,x_j)\rightarrow K_{ij}$ such that $K\succeq 0$.  Table \ref{table:kernel_defs} lists several classic kernel functions used in text classification, each corresponding to a different implicit mapping to feature space.
\begin{table}[h!]
\begin{center}
\extrarowheight 0.8ex
\begin{tabular}{r|l}
Linear kernel & $k(x_i,x_j)=\langle x_i,x_j\rangle$ \\ 
Gaussian kernel & $k(x_i,x_j)=e^{-\|x_i-x_j\|^2/\sigma}$\\ 
Polynomial kernel & $k(x_i,x_j)=(\langle x_i,x_j\rangle + 1)^d$\\ 
Bag-of-words kernel & $k(x_i,x_j)=\frac{\langle x_i,x_j\rangle}{\|x_i\|\|x_j\|}$\\ 
\end{tabular}
\end{center}
\caption{Several classic kernel functions.}
\label{table:kernel_defs}
\end{table}

Many efficient algorithms have been developed for solving the quadratic program (\ref{eq:svm_dual}).  A common technique uses sequential minimal optimization (SMO), which is coordinate descent where all but two variables are fixed and the remaining two-dimensional problem is solved explicitly.  All experiments in this paper use the LIBSVM \cite{Chan2001} package implementing this method.

\subsection{Data}
Data vectors $x_i$ in the following experiments are formed using text features and equity returns features.  Text features are extracted from press releases as a \emph{bag-of-words}.  A fixed set of important words referred to as the dictionary is predetermined; in this instance, 619 words such as \emph{increase}, \emph{decrease}, \emph{acqui}, \emph{lead}, \emph{up}, \emph{down}, \emph{bankrupt}, \emph{powerful}, \emph{potential}, and \emph{integrat} are considered.  Stems of words are used so that words such as \emph{acquired} and \emph{acquisition} are considered identical. We use the following \emph{Microsoft} press release and its bag-of-words representation in Figure \ref{fig:text_example} as an example. Here, $x_{ij}$ is the number of times that the $j^{th}$ word in the dictionary occurs in the $i^{th}$ press release.

\begin{figure}[h!]\small{
\emph{LONDON — Dec. 12, 2007 — Microsoft Corp. has \textbf{acquired} Multimap, one of the United Kingdom's top 100 technology companies and one of the \textbf{leading} online mapping services in the world. The \textbf{acquisition} gives Microsoft a \textbf{powerful} new location and mapping technology to complement existing offerings such as Virtual Earth, Live Search, Windows Live services, MSN and the aQuantive advertising platform, with future \textbf{integration} \textbf{potential} for a range of other Microsoft products and platforms. Terms of the deal were not disclosed.}
\begin{center}
\begin{tabular}{|c|c|c|c|c|c|c|c|c|c|}
\hline
increas&decreas&acqui&lead&up&down&bankrupt&powerful&potential&integrat\\
\hline
0&0&2&1&0&0&0&1&1&1\\
\hline
\end{tabular}
\end{center}
}
\caption{Example of \emph{Microsoft} press release and the corresponding bag-of-words representation.  Note that words in the dictionary are stems.}
\label{fig:text_example}
\end{figure}

These numbers are transformed using term frequency-inverse document frequency weighting (tf-idf) defined by
\BEQ \BA{ll}
\mbox{TF-IDF}(i,j)=\mbox{TF}(i,j)\cdot\mbox{IDF}(i),&\mbox{IDF}(i)=\log\frac{N}{\mbox{DF}(i)}
\EA
\label{eq:tfidf} \EEQ
where $\mbox{TF}(i,j)$ is the number of times that term $i$ occurs in document $j$ (normalized by the number of words in document $j$) and $\mbox{DF}(i)$ is the number of documents in which term $i$ appears.  This weighting increases the importance of words that show up often within a document but also decreases the importance of terms that appear in too many documents because they are not useful for discrimination.  Other advanced text representations include latent semantic analysis \cite{Deer90}, probabilistic latent semantic analysis \cite{Hofm01}, and latent dirichlet allocation \cite{Blei2003}.  In regards to equity return features, $x_i$ corresponds to a time series of 5 returns (taken at 5 minute intervals and calculated with 15 minute lags) based on equity prices leading up to the time when the press release is published.  Press releases published before 10:10 am thus do not have sufficient stock price data to create the equity returns features used here and most experiments will only consider news published after 10:10 am.

Experiments are based on press releases issued during the eight year period 2000-2007 by \emph{PRNewswire}. We focus on news related to publicly traded companies that issued at least 500 press releases through \emph{PRNewswire} in this time frame.  Press releases tagged with multiple stock tickers are discarded from experiments.  Intraday price data is taken from the \emph{NYSE Trade and Quote Database (TAQ)} through \emph{Wharton Research Data Services}.

The eight year horizon is divided into monthly data.  In order to simulate a practical environment, all decision models are calibrated on one year of press release data and used to make predictions on articles released in the following month; thus all tests are out-of-sample.  After making predictions on a particular month, the one year training window slides forward by one month as does the one month test window.

\begin{figure}[h!] \begin{center}
  \begin{tabular} {cc}
     \psfrag{title}[b]{\small{Aggregated Testing Data}}
     \psfrag{num}[b]{\small{Number Press Releases}}
     \psfrag{min}[t]{\small{Minutes}}
     \includegraphics[width=0.49 \textwidth]{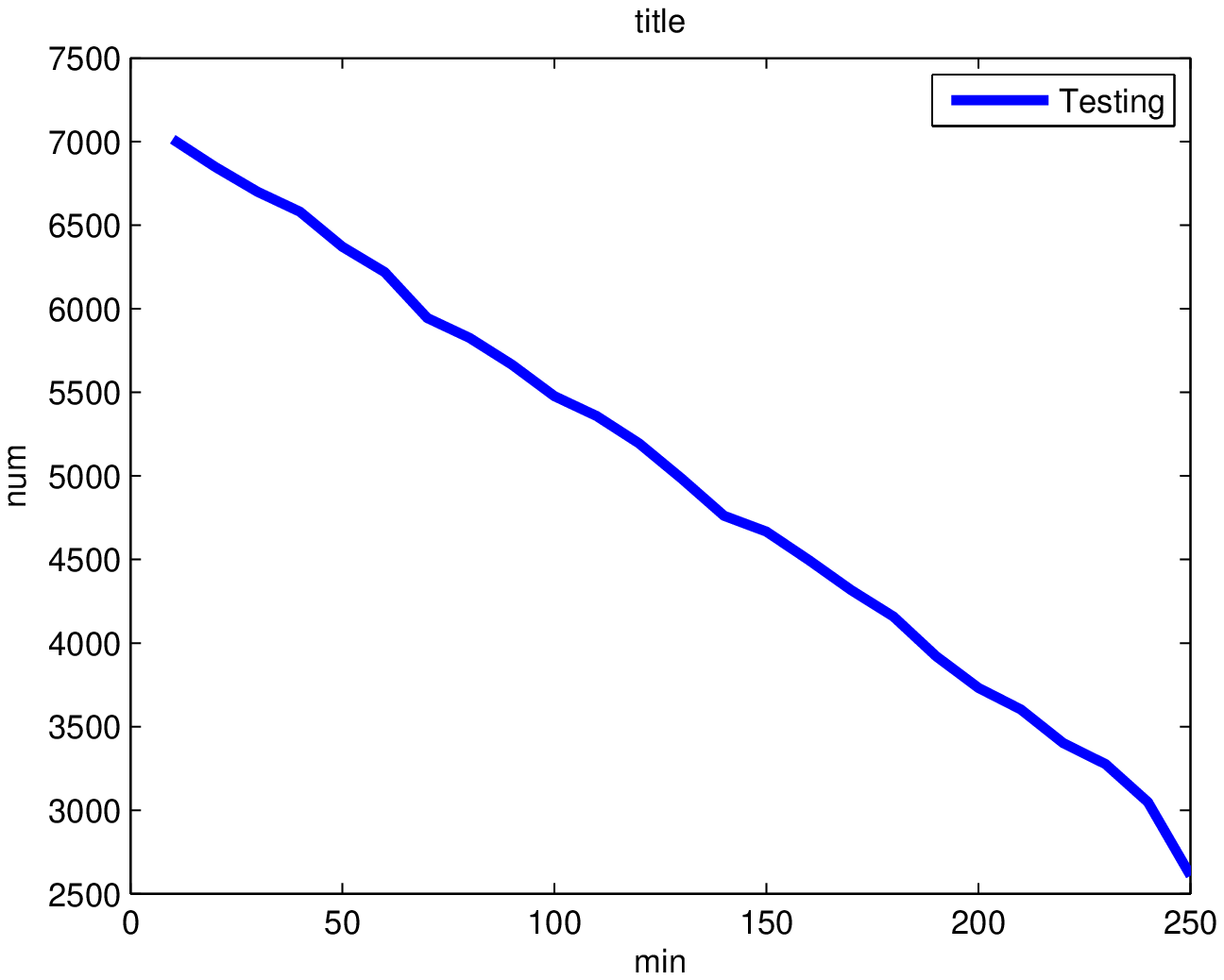}&
    \psfrag{title}[b]{\small{Avg. Training/Testing Data Per Window}}
    \psfrag{num}[b]{\small{Number Press Releases}}
    \psfrag{min}[t]{\small{Minutes}}
    \includegraphics[width=0.49 \textwidth]{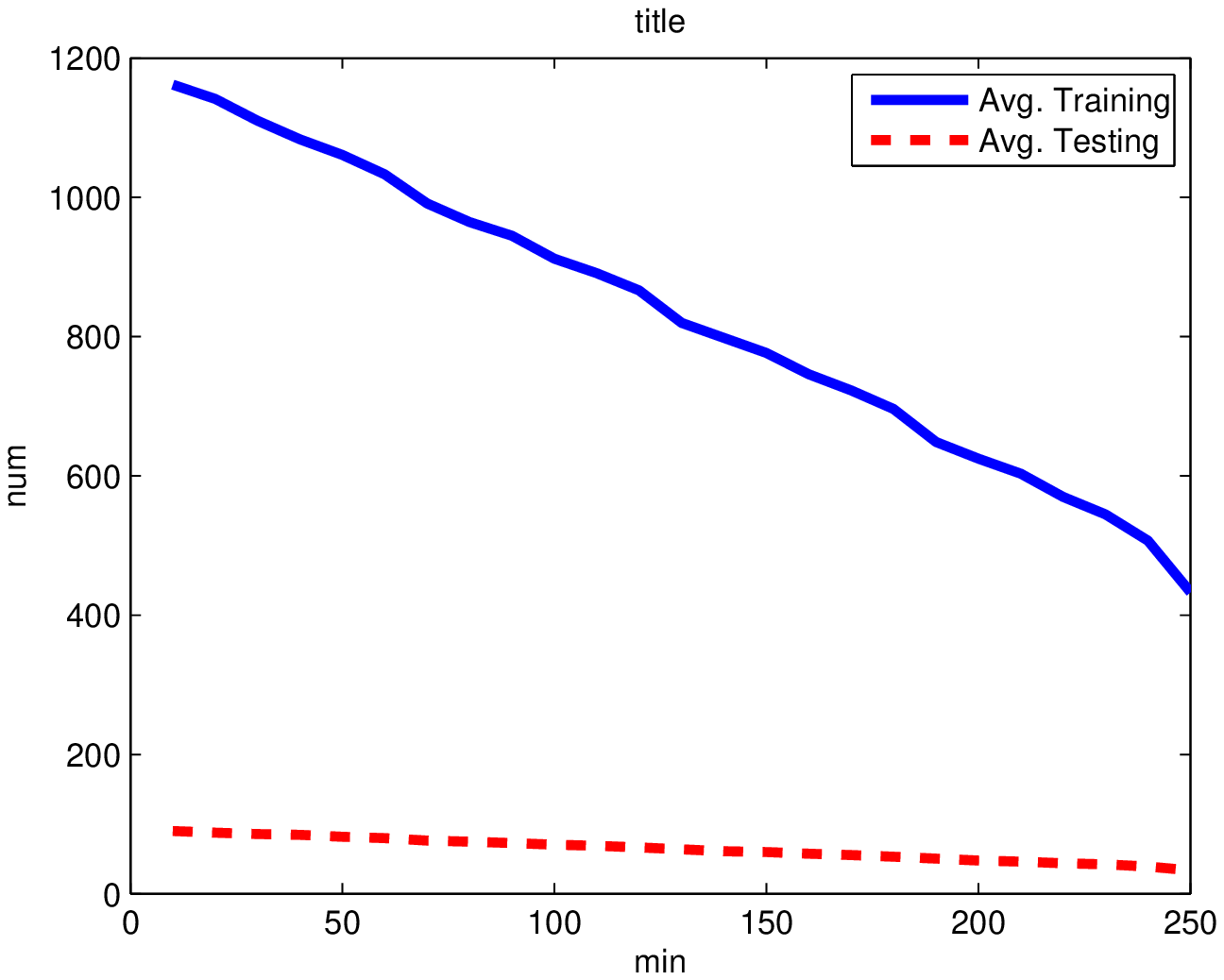}
  \end{tabular}
\caption{Aggregate (over all windows) amount of test press releases (left) and average training/testing set per window (right).  Average training and testing windows are one year and one month, respectively.  Aggregated test data over all windows is used to calculate all performance measures.}
\label{fig:DataSize}
\end{center} \end{figure}

Price data is used for each press release for a fixed period prior to the release and at each 10 minute interval following the release of the article up to 250 minutes. When, for example, news is released at 3 pm, price data exists only for 60 minutes following the news (because the business day ends at 4 pm), so this particular article is discarded from experiments that make predictions with time horizons longer than 60 minutes.  Overall, this means that training and testing data sizes decrease with the forecasting horizon.  Figure \ref{fig:DataSize} displays the overall amount of testing data (left) and the average amount of training and testing data used in each time window (right).

\subsection{Performance Measures}
Most kernel functions in Table \ref{table:kernel_defs} contain parameters requiring calibration.  A set of reasonable values for each parameter is chosen, and for each combination of parameter values, we perform $n$-fold cross-validation to optimize parameter values.  Training data is separated into $n$ equal folds.  Each fold is pulled out successively, and a model is trained on the remaining data and tested on the extracted fold.  A predefined classification performance measure is averaged over the $n$ test folds and the optimal set of parameters is determined as those that give the best performance.  Since the distribution of words occurring in press releases may change over time, we perform chronological one-fold cross validation here.  Training data is ordered according to release dates, after which a model is trained on all news published before a fixed date and tested on the remaining press releases (the single fold).  Several potential measures are defined in Table \ref{table:perf}.  Note that the SVM Problem (\ref{eq:svm_dual}) also has a parameter $C$ that must be calibrated using cross-validation.

\begin{table}[h!]
\begin{center}
\extrarowheight 0.8ex
\begin{tabular}{|r|c|}
\hline
Annualized sharpe ratio: & $\sqrt{T}\frac{E[r]}{\sigma}$ \\ \hline
Accuracy: & $\frac{TP+TN}{TP+TN+FP+FN}$\\ \hline
Recall: & $\frac{TP}{TP+FN}$\\ \hline
\end{tabular}
\end{center}
\caption{Performance measures. $T$ is the number of periods per year (12 for monthly, 252 for daily).  $E[r]$ is the expected return per period of a given trading strategy, and $\sigma$ is the standard deviation of~$r$. For binary classification, $TP$, $TN$, $FP$, and $FN$ are, respectively, true positives, true negatives, false positives, and false negatives.\label{table:perf}}
\end{table}

Beyond standard accuracy and recall measures, we measure prediction performance with a more financially intuitive metric, the Sharpe ratio, defined here as the ratio of the expected return to the standard deviation of returns, for the following (fictitious) trading strategy: every time a news article is released, a bet is made on the stock return and we either win or lose \$1 according to whether or not the prediction is correct.  Daily returns are computed as the return of playing this game on each press release published on a given day.  The Sharpe ratio is estimated using the mean and standard deviation of these daily returns, then annualized. Additional results are given using the classic performance measure: accuracy, defined as the percentage of correct predictions made, however all results are based on cross-validating over Sharpe ratios.  Accuracy is displayed due to its intuitive meaning in binary classification, but it has no direct financial interpretation. Another potential measure is recall, defined as the percentage of positive data points that are predicted positive. In general, a tradeoff between accuracy and recall would be used as a measure in cross-validation. Here instead, we tradeoff risk versus returns by optimizing the Sharpe ratio.

\subsection{Predicting equity movements with text \emph{or} returns}
\label{subsec:pred_with_svm}
Support vector machines are used here to make predictions on stock returns when news regarding the company is published. In this section, the input feature vector to SVM is either a bag-of-words text vector or a time series of past equity returns, as SVM only inputs a single feature vector.  Predictions are considered at every 10 minute interval following the release of an article up to either a maximum of 250 minutes or the close of the business day; i.e. if the article comes out at 10:30 am, we make predictions on the equity returns at 10:40 am, 10:50 am, ... , until 2:40 pm.  Only articles released during the business day are considered here.

Two different classification tasks are performed.  In one experiment, the direction of returns is predicted by labeling press releases according to whether the future return is positive or negative.  In the other experiment, we predict abnormal returns, defined as an absolute return greater than a predefined threshold.  Different thresholds correspond to different classification tasks and we expect larger jumps to be easier to predict than smaller ones because the latter may not correspond to true abnormal returns.  This will be verified in experiments below.

\begin{figure}[h!] \begin{center}
  \begin{tabular} {cc}
     \psfrag{title}[b]{\small{Accuracy using Returns}}
     \psfrag{acc}[b]{\small{Accuracy}}
     \psfrag{min}[t]{\small{Minutes}}
     \includegraphics[width=0.49 \textwidth]{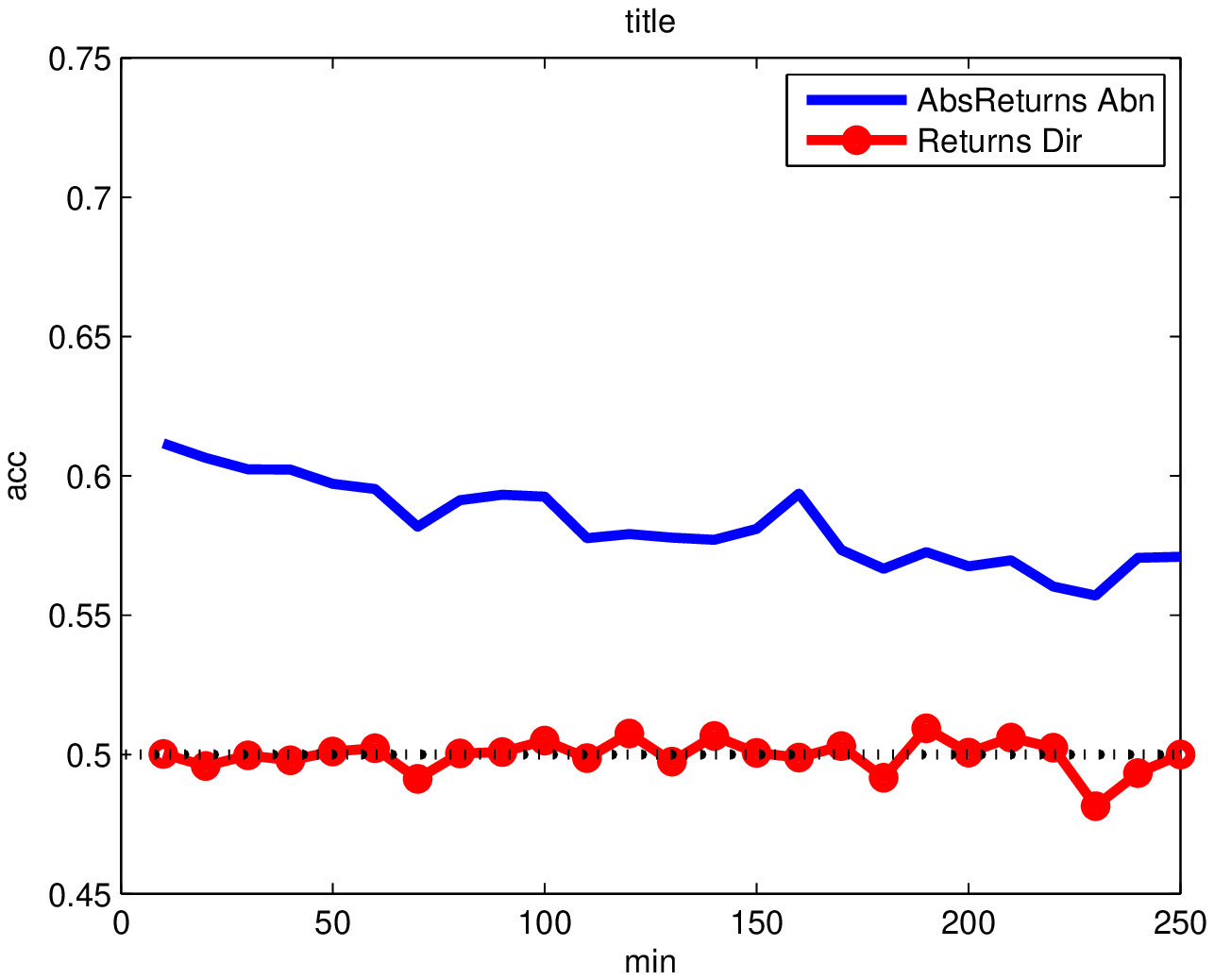}&
     \psfrag{title}[b]{\small{Accuracy using Text}}
     \psfrag{acc}[b]{\small{Accuracy}}
     \psfrag{min}[t]{\small{Minutes}}
     \includegraphics[width=0.49 \textwidth]{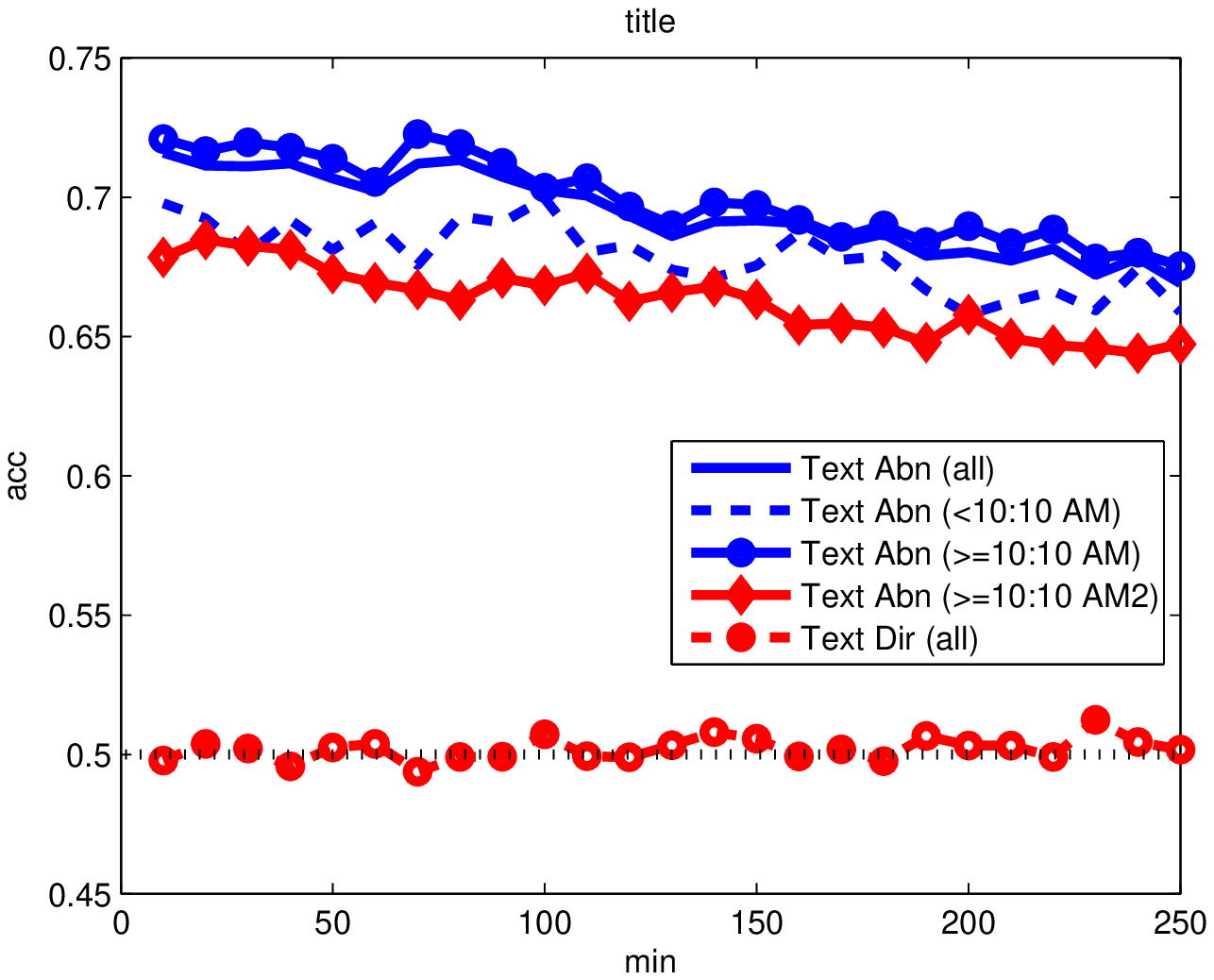}\\\\
    \psfrag{title}[b]{\small{Sharpe Ratio using Returns}}
    \psfrag{sharpe}[b]{\small{Sharpe Ratio}}
    \psfrag{min}[t]{\small{Minutes}}
    \includegraphics[width=0.49 \textwidth]{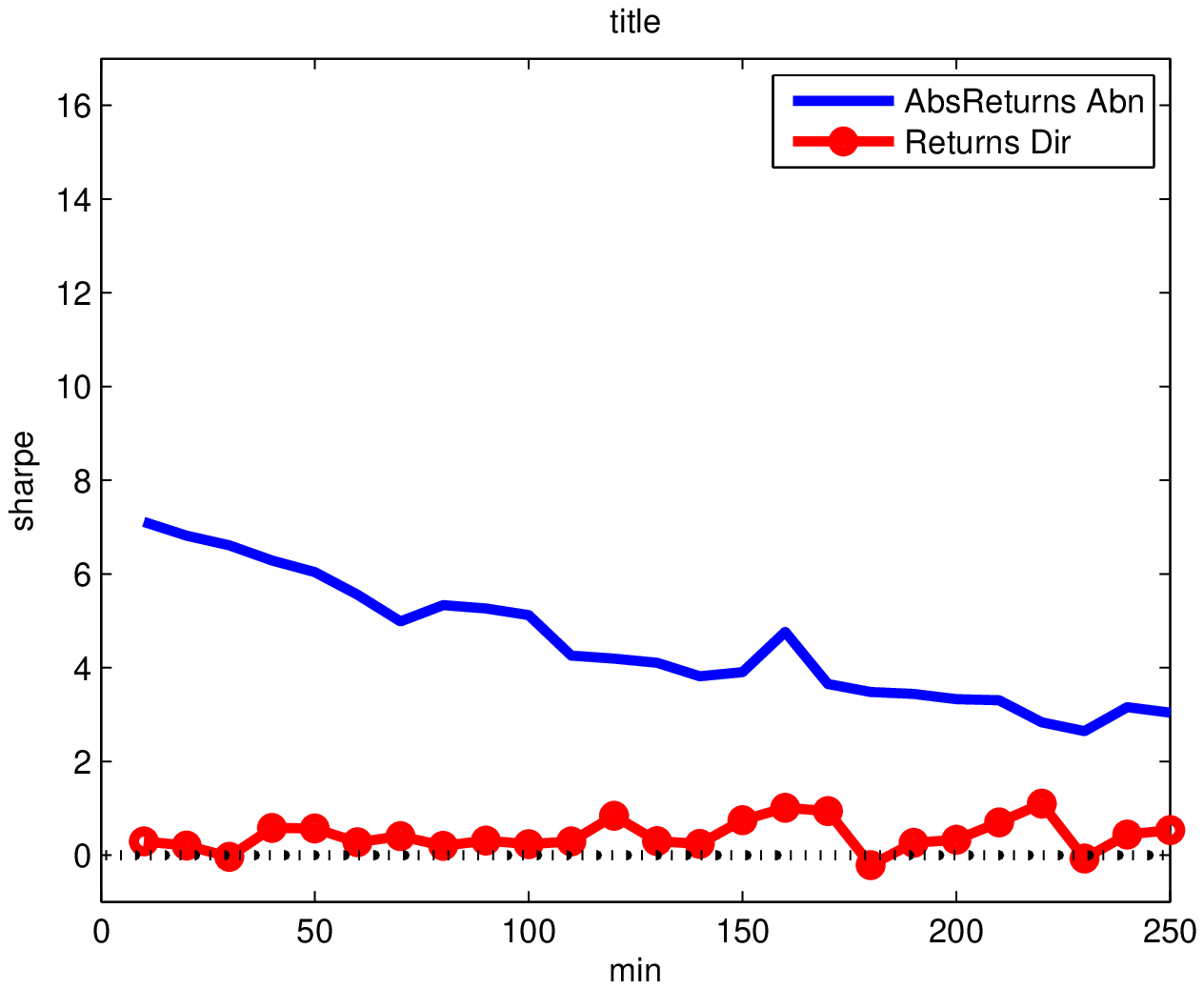}&
    \psfrag{title}[b]{\small{Sharpe Ratio using Text}}
    \psfrag{sharpe}[b]{\small{Sharpe Ratio}}
    \psfrag{min}[t]{\small{Minutes}}
    \includegraphics[width=0.49 \textwidth]{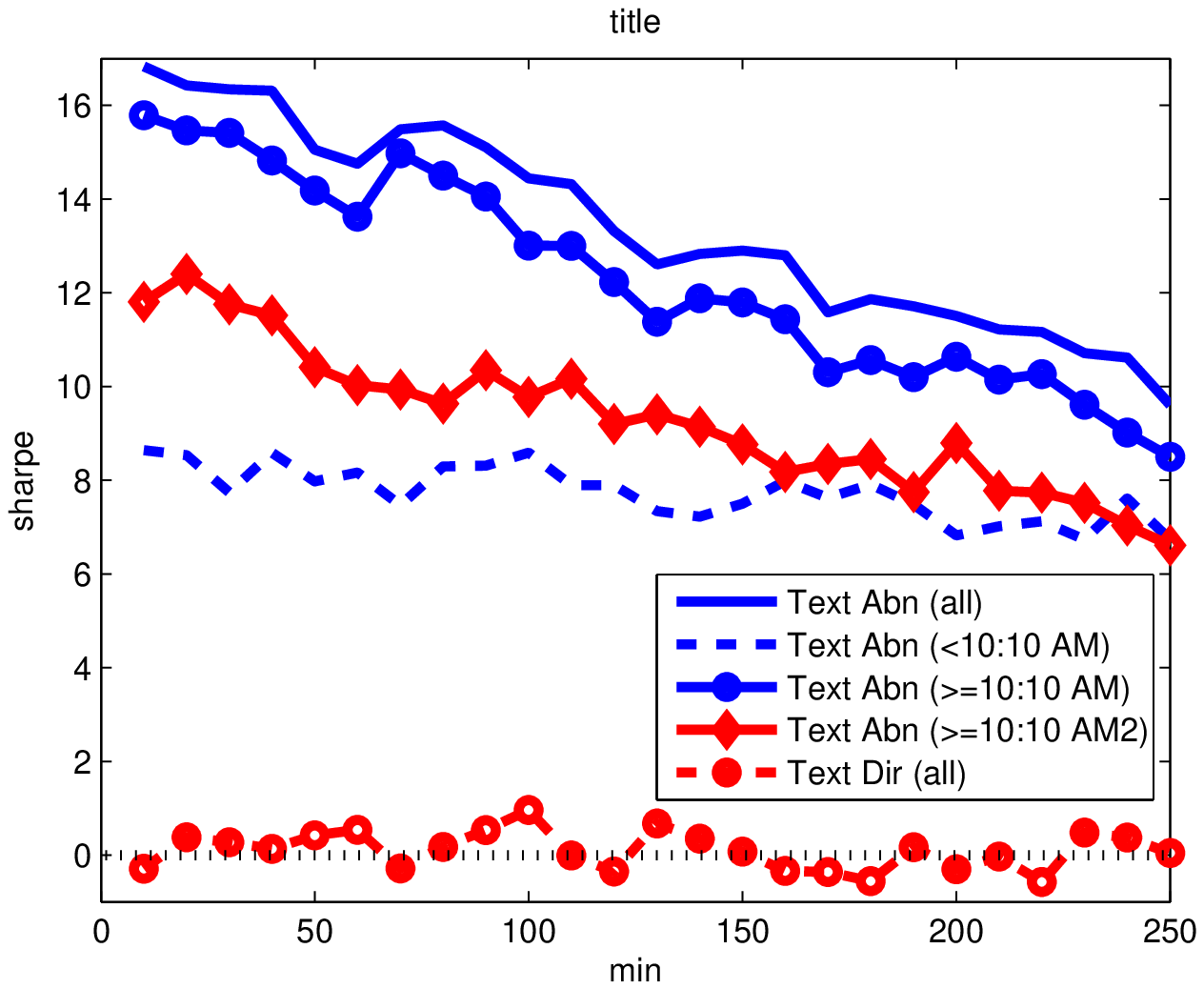}
  \end{tabular}
\caption{Accuracy and annualized daily Sharpe ratio for predicting abnormal returns (Abn) or direction of returns (Dir) using returns and text data with linear kernels.  Performance using text is given for both the full data set as well as the reduced data set that is used for experiments with returns features.  The curves labeled with $\ge$10:10 AM trains models using the complete data set including articles released at the open of the business day while the curved labeled with $\ge$10:10 AM2 does not use the first 40 minutes of news to train models.  Each point \emph{z} on the x-axis corresponds to predicting an abnormal return \emph{z} minutes after each press release is issued.  The $75^{th}$ percentile of absolute returns observed in the training data is used as the threshold for defining an abnormal return.}
\label{fig:single_kernels_75}
\end{center} \end{figure}

Performance of predicting the direction of equity returns following press releases is displayed in Figure~\ref{fig:single_kernels_75} and shows the weakest performance, using either a time series of returns (left) or text (right) as features.  No predictability is found in the direction of equity returns (since the Sharpe ratio is near zero and the accuracy remains close to 50\%).  This is consistent with literature regarding stock return predictability.  All results displayed here use linear kernels with a single feature type.  Instead of the fictitious trading strategy used for abnormal return predictions, directional results use a buy and sell (or sell and buy) strategy based on the true equity returns.  Similar performance using gaussian kernels was observed in independent experiments.

While predicting direction of returns is a difficult task, abnormal returns appear to be predictable using either a time series of absolute returns or the text of press releases.  Figure \ref{fig:single_kernels_75} shows that a time series of absolute returns contains useful information for intraday predictions (left), while even better predictions can be made using text (right).  The threshold for defining abnormal returns in each window is the $75^{th}$ percentile of absolute returns observed in the training data.

As described above, experiments with returns features only use news published after 10:10 am.  Thus, performance using text kernels is given for both the full data set with all press released during the business day as well as the reduced data set to compare against experiments with returns features.  Performance from the full data set is also broken down according to press released before and after 10:10 am. The difference between the curves labeled $\ge$10:10 AM and $\ge$10:10 AM2 is that the former trains models using the complete data set including articles released at the open of the business day while the latter does not use the first 40 minutes of news to train models.  The difference in performance might be attributed to the importance of these articles.  The Sharpe ratio using the reduced data set is greater than that for news published before 10:10 am because fewer articles are published in the first 40 minutes than are published during the remainder of the business day.

Note that these very high Sharpe ratio are most likely due to the simple \emph{strategy} that is traded here; this does not imply that such a high Sharpe ratio can be generated in practice but rather indicates a potential statistical arbitrage. The decreasing trend observed in all performance measures over the intraday time horizon is intuitive: public information is absorbed into prices over time, hence articles slowly lose their predictive power as the prediction horizon increases.

\begin{figure}[h!] \begin{center}
  \begin{tabular} {cc}
     \psfrag{title}[b]{\small{Accuracy using Returns/Text}}
     \psfrag{acc}[b]{\small{Accuracy}}
     \psfrag{min}[t]{\small{Minutes}}
     \includegraphics[width=0.49 \textwidth]{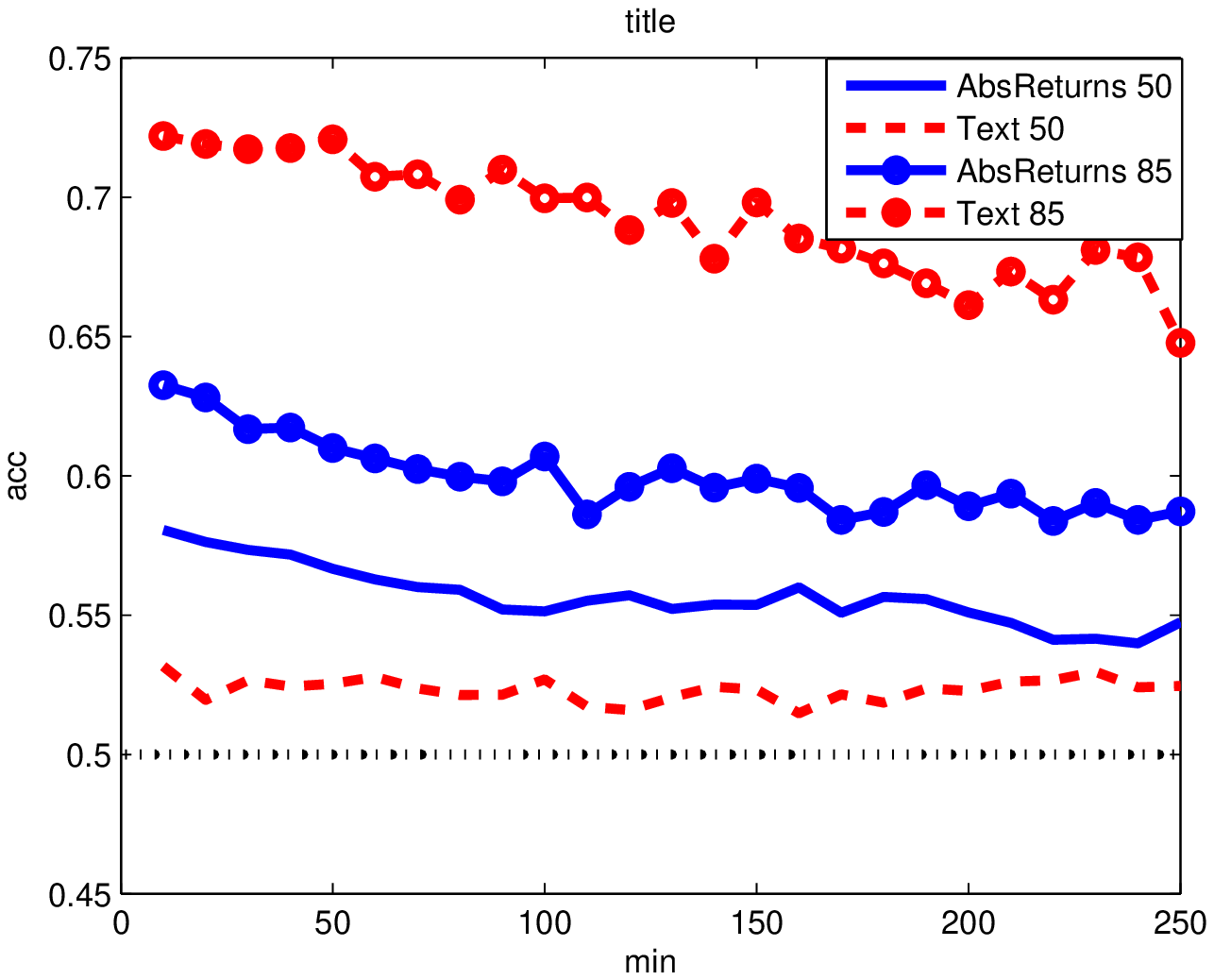}&
     \psfrag{title}[b]{\small{Sharpe Ratio using Returns/Text}}
     \psfrag{sharpe}[b]{\small{Sharpe Ratio}}
     \psfrag{min}[t]{\small{Minutes}}
     \includegraphics[width=0.49 \textwidth]{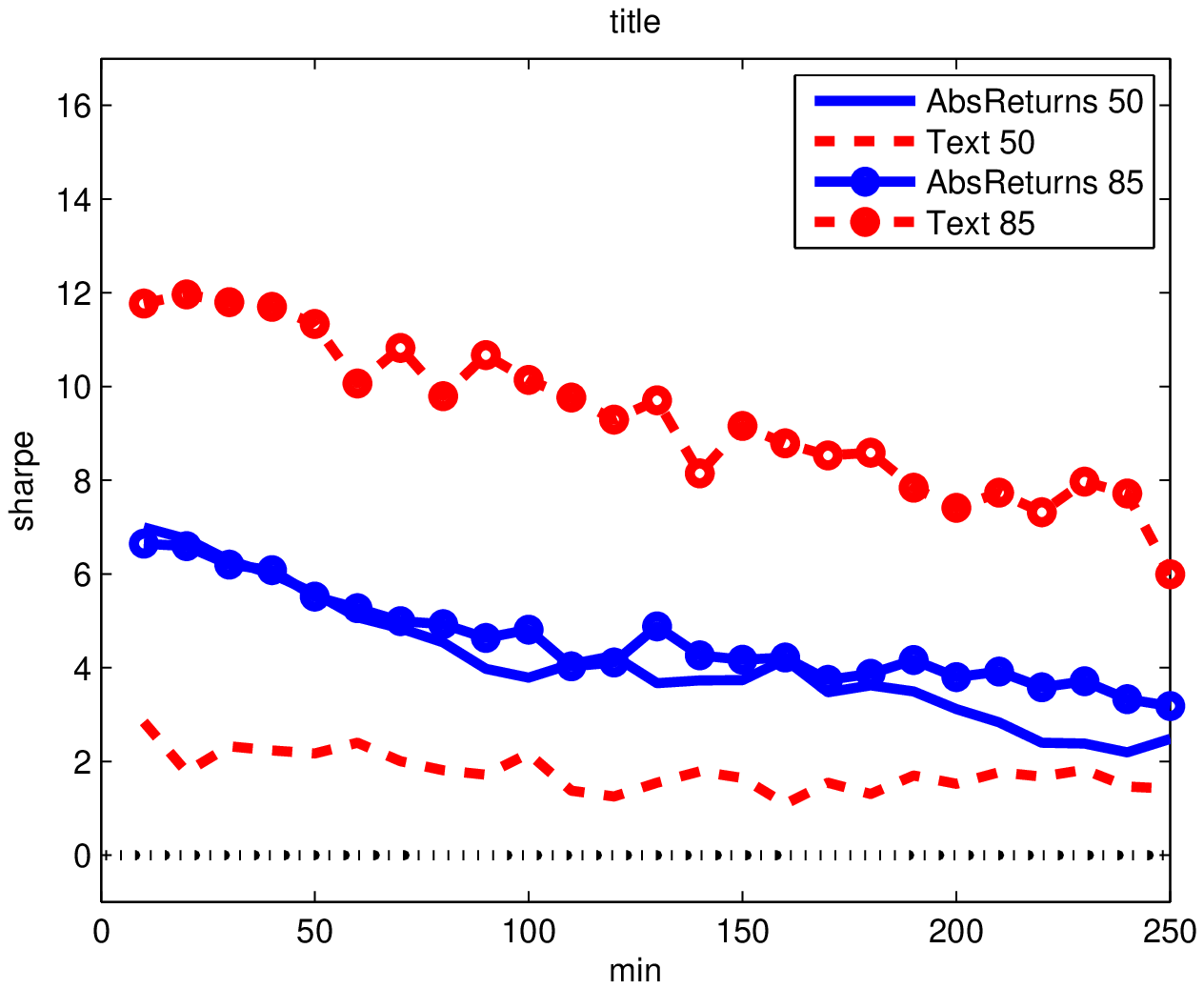}
  \end{tabular}
\caption{Accuracy and annualized daily Sharpe ratio for predicting abnormal returns using returns and text data with linear kernels.  Each point \emph{z} on the x-axis corresponds to predicting an abnormal return \emph{z} minutes after each press release is issued.  The $50^{th}$ and $85^{th}$ percentile of absolute returns observed in the training data are used as thresholds for defining abnormal returns.}
\label{fig:single_kernels_movement_50_85}
\end{center} \end{figure}

\begin{figure}[h!] \begin{center}
  \begin{tabular} {cc}
     \psfrag{title}[b]{\small{Accuracy using Returns/Text}}
     \psfrag{acc}[b]{\small{Accuracy}}
     \psfrag{perc}[t]{\small{Threshold Percentile}}
     \includegraphics[width=0.49 \textwidth]{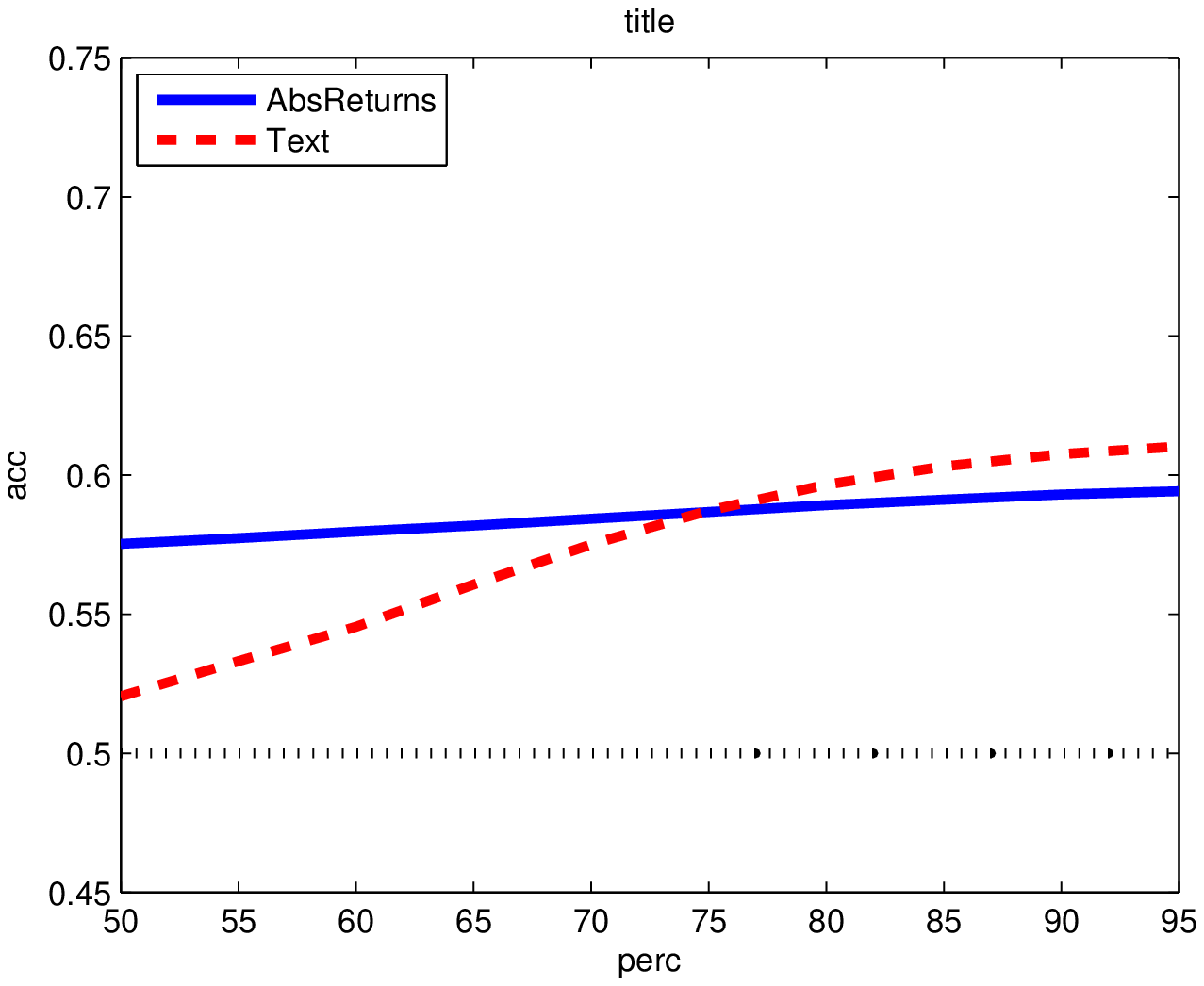}&
     \psfrag{title}[b]{\small{Sharpe Ratio using Returns/Text}}
     \psfrag{sharpe}[b]{\small{Sharpe Ratio}}
     \psfrag{perc}[t]{\small{Threshold Percentile}}
     \includegraphics[width=0.49 \textwidth]{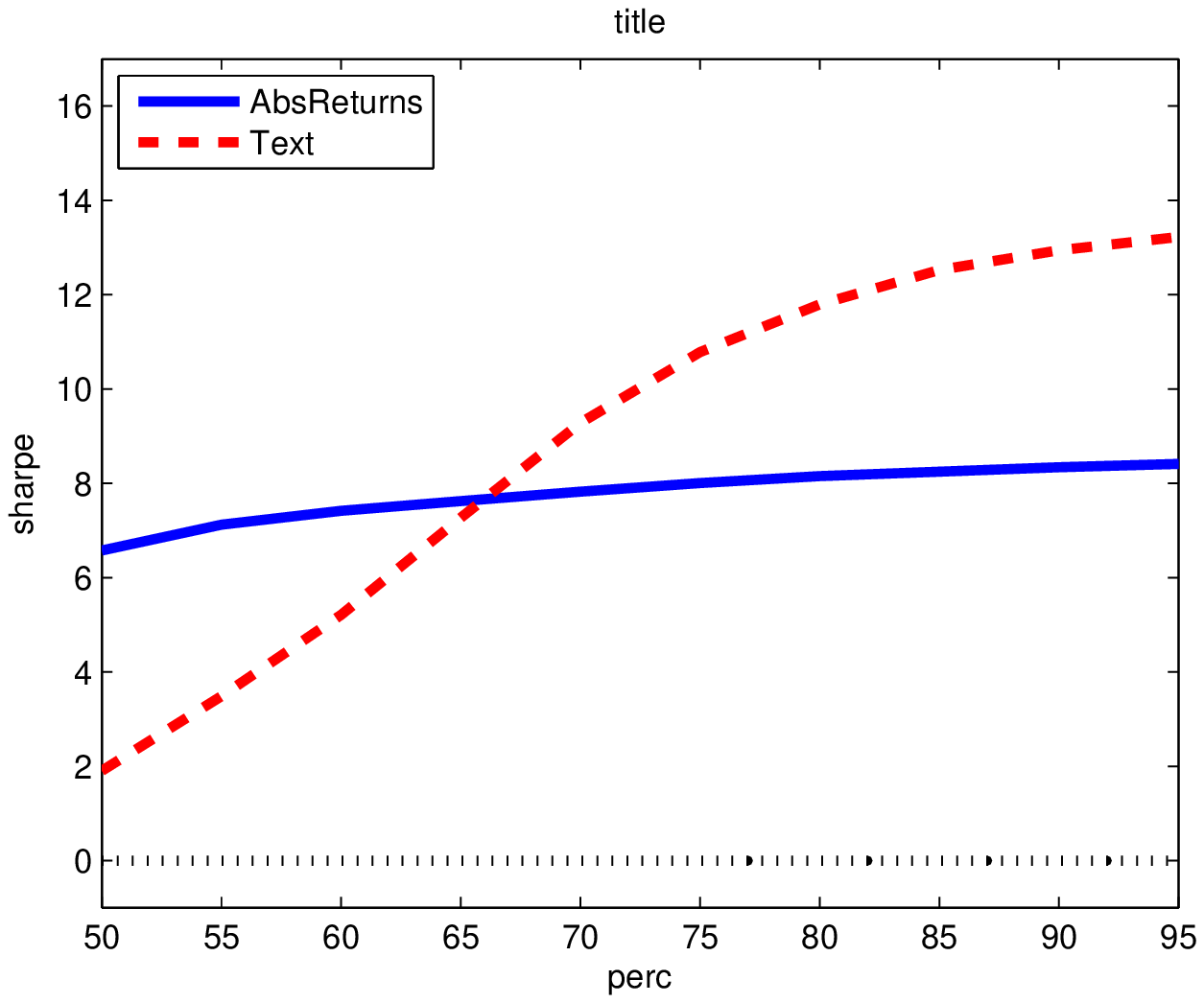}
  \end{tabular}
\caption{Accuracy and annualized sharpe ratio for predicting abnormal returns 20 minutes into the future as the percentile for thresholds is increased from 50\% to 95\%.  Linear kernels with absolute returns and text are used.  For 25-35\% of press releases, news has a bigger impact on future returns than past market data.}
\label{fig:single_kernels_movements_perc_vs_perf_20mins}
\end{center} \end{figure}

Figure \ref{fig:single_kernels_movement_50_85} compares performance in predicting abnormal returns when the threshold is taken at either the $50^{th}$ or $85^{th}$ percentile of absolute returns within the training set.  Results using linear kernels and annualized Sharpe ratios (using daily returns) are shown here.  Decreasing the threshold to the $50^{th}$ percentile slightly decreases performance when using absolute returns.  However, there is a huge decrease in performance when using text.  Increasing the threshold to the $85^{th}$ percentile improves performance relative to the $75^{th}$ percentile in all measures.  This demonstrates the sensitivity of performance with respect to this threshold.  The $50^{th}$ percentile of absolute returns from the data set is not large enough to define a \emph{true} abnormal return, whereas the $75^{th}$ and $85^{th}$ percentiles do define abnormal jumps.  Absolute returns are known to have predictability for small movements, but the question remains as to why text is a poor source of information for predicting small jumps.  Figure \ref{fig:single_kernels_movements_perc_vs_perf_20mins} illustrates the impact of this percentile threshold on performance. Predictions are made 20 minutes into the future.  For 25-35\% of press releases, news has a bigger impact on future returns than past market data.

\subsection{Time of Day Effect}
\label{subsec:time of day}  
\begin{figure}[h!] \begin{center}
    \psfrag{title}[b]{\footnotesize{Average Absolute (10 minute) Returns following Press Releases}}
     \psfrag{absret}[b]{\footnotesize{Average Absolute Return}}
    \includegraphics[width=\textwidth]{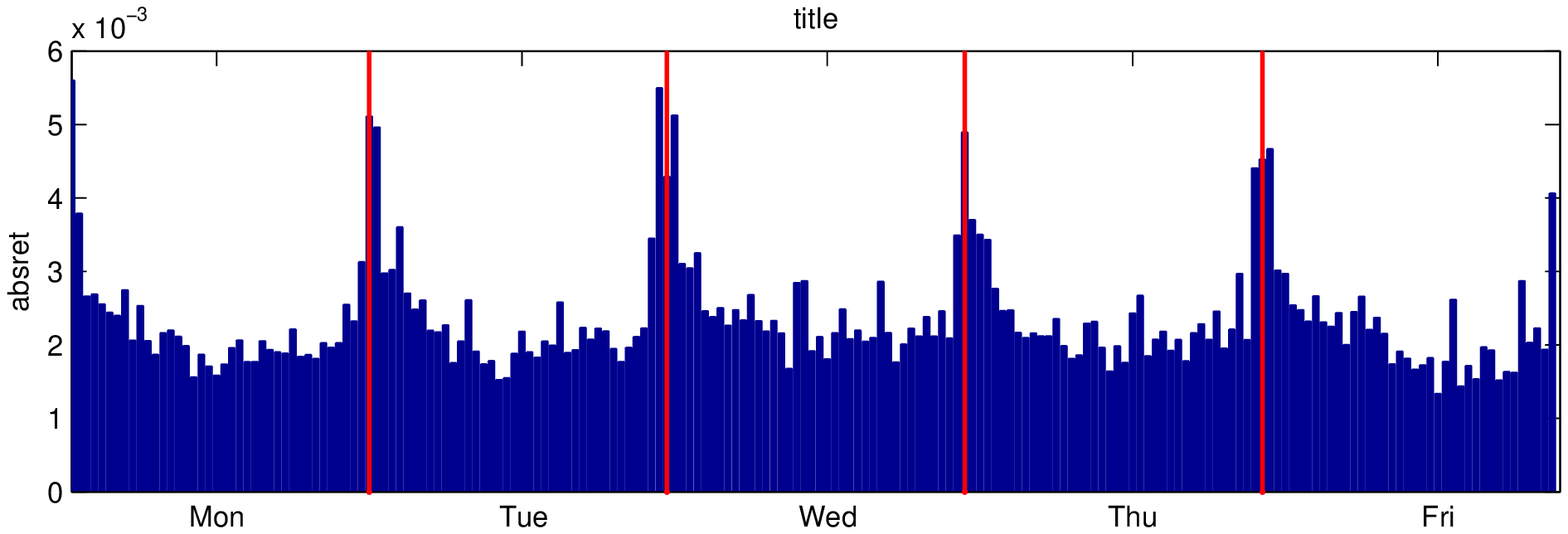}
\caption{Average absolute (10 minute) returns following press released during the business day. Red lines are drawn between business days.}
\label{fig:timeofday}
\end{center} \end{figure}

Other publicly available information aside from returns and text should be considered when predicting movements of equity returns. The time of day has a strong impact on absolute returns, as demonstrated by \citeasnoun{Ande97} for the S\&P 500.  Figure \ref{fig:timeofday} shows the time of day effect following the release of press from the \emph{PRNewswire} data set.  It is clear that absolute returns following press released early (and late) in the day are on average much higher than during midday.

\begin{figure}[h!] \begin{center}
  \begin{tabular} {cc}
     \psfrag{title}[b]{\small{Accuracy using Time of Day}}
     \psfrag{acc}[b]{\small{Accuracy}}
     \psfrag{min}[t]{\small{Minutes}}
     \includegraphics[width=0.49 \textwidth]{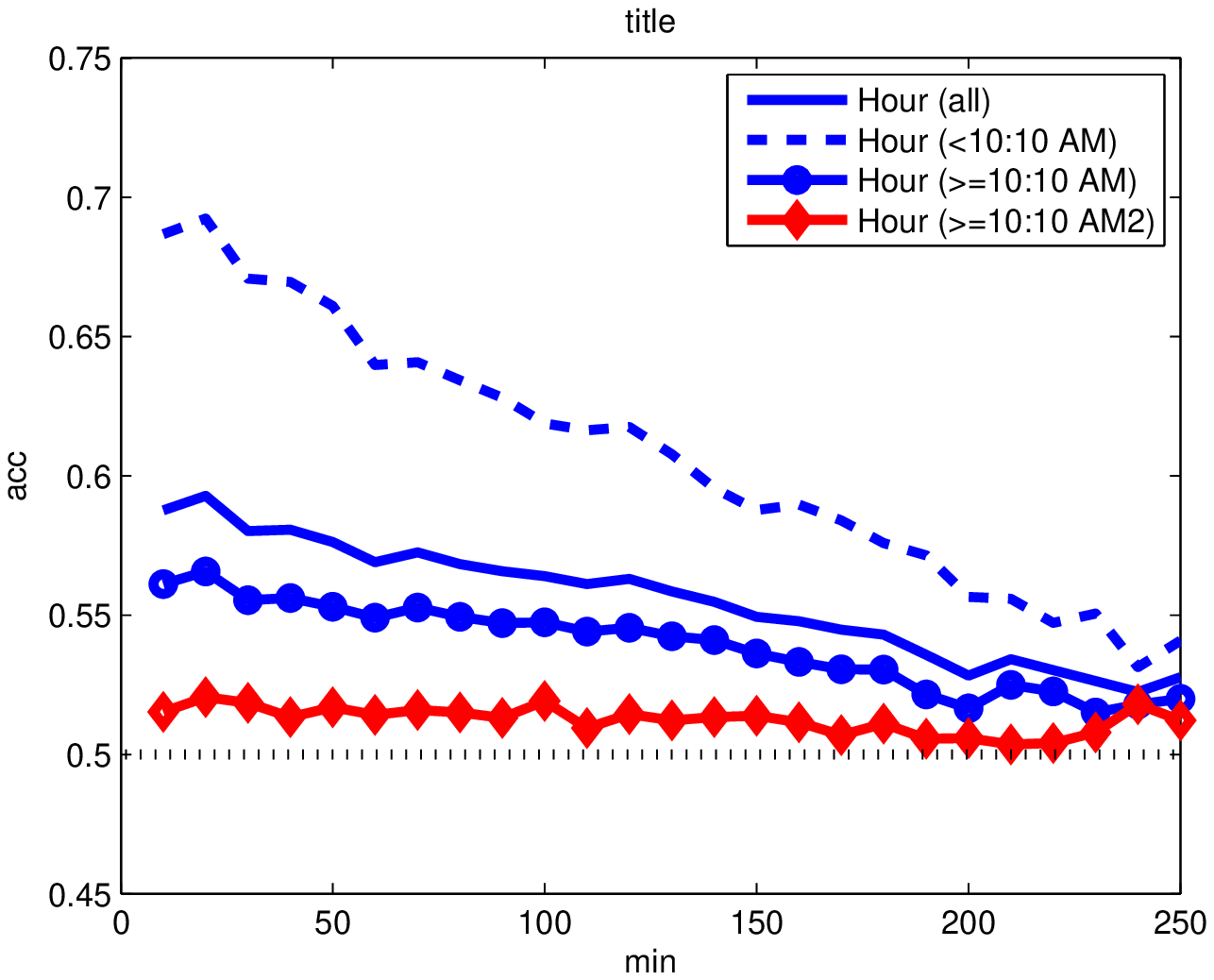}&
     \psfrag{title}[b]{\small{Sharpe Ratio using Time of Day}}
     \psfrag{sharpe}[b]{\small{Sharpe Ratio}}
     \psfrag{min}[t]{\small{Minutes}}
     \includegraphics[width=0.49 \textwidth]{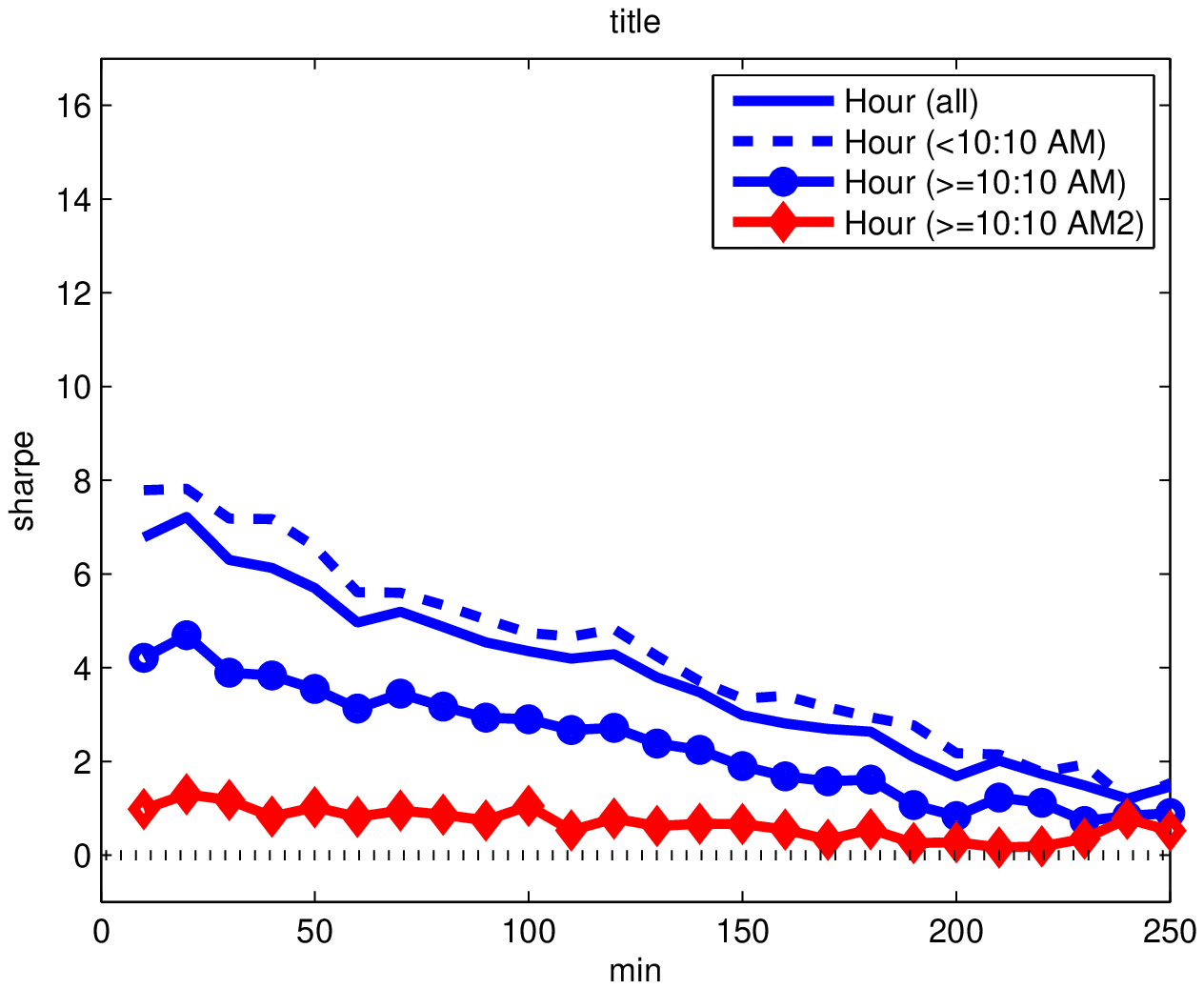}
  \end{tabular}
\caption{Accuracy and annualized daily sharpe ratio for predicting abnormal returns using time of day.  Performance using time of day is given for both the full data set as well as the reduced data set that is used for experiments with returns features.  The curves labeled with $\ge$10:10 AM trains models using the complete news data set including articles released at the open of the business day while the curved labeled with $\ge$10:10 AM2 does not use the first 40 minutes of news to train models.  Each point \emph{z} on the x-axis corresponds to predicting an abnormal return \emph{z} minutes after each press release is issued.  The $75^{th}$ percentile of absolute returns observed in the training data are used as thresholds for defining abnormal returns.}
\label{fig:single_kernels_hour_75}
\end{center} \end{figure}

We use the time stamp of the press release as a feature for making the same predictions as above.  A binary feature vector $x\in\reals^3$ is created to label each press release as published before 10:30 am, after 3 pm, or in between.  Linear kernels are created from these features and used in SVM for the same experiments as above with absolute returns and text features and results are displayed in Figure \ref{fig:single_kernels_hour_75}.  Note that gaussian kernels have exactly the same performance when using these binary features.  As was done for the analysis with text data, performance is shown when using all press released during the business day as well as the reduced data set with news only published after 10:10 am (labels are the same as were described for text).  Training SVM with data from the beginning of the day is clearly important since the curve labeled $\ge$10:10 AM2 has the weakest performance.

The improved performance of the curve labeled $\ge$10:10 AM over $\ge$10:10 AM2 can be attributed to the pattern seen in Figure \ref{fig:timeofday}.  Training with the full data set allows the model to distinguish between absolute returns early in the day versus midday.  Similar experiments using day of the week features showed very weak performance and are thus not displayed.  While the time of day effect exhibits predictability, note that the experiments with text and absolute returns data do not use any time stamp features and hence performance with text and absolute returns should not be attributed to any time of day effects.  Furthermore, experiments below for combining the different pieces of publicly available information will show that these time of day effects are less useful than the text and returns data.  There are of course other related market microstructure effects that could be useful for predictability, such as the amount of news released throughout the day or the industry of the respective companies.

\subsection{Predicting daily equity movements and trading covered call options}
\label{subsec:daily_movements}
While the main focus is intraday movements, we next use text and absolute returns to make daily predictions on abnormal returns and show how one can trade on these predictions.  These experiments use the same text data as above for a subset of 101 companies (daily options data was not obtained for all companies).  Returns data is also an intraday time series as above, but is here computed as the 5, 10, ..., 25 minute return prior to press releases.  Daily equity data is obtained from the \emph{YAHOO!Finance} website and the options data is obtained using \emph{OptionMetrics} through \emph{Wharton Research Data Services}.

Rather than the fictitious trading strategy above, delta-hedged covered call options are used to bet on abnormal returns (intraday options data was not available hence the use of a fictitious strategy above).  In order to bet on the occurrence of an abnormal return, the strategy takes a long position in a call option, and, since the bet is not on the direction of the price movement, the position is kept delta neutral by taking a short position in delta shares of stock (delta is defined as the change in call option price resulting from a \$1 increase in stock price, here taken from the \emph{OptionMetrics} data).  The position is exited the following day by going short the call option and long delta shares of stock.  A bet against an abnormal return takes the opposite positions.  Equity positions use the closing prices following the release of press and the closing price the following day.  Option prices (buy and sell) use an average of the highest closing bid and lowest closing ask price observed on the day of the press release.  To normalize the size of positions, we always take a position in delta times \$100 worth of the respective stock and the proper amount of the call option.

\begin{figure}[h!] \begin{center}
  \begin{tabular} {cc}
     \psfrag{title}[b]{\small{P\&L of Predicting Abnormal Returns}}
     \psfrag{ylab}[b]{\small{Profit and Loss}}
     \psfrag{xlab}[t]{\small{Change in Stock Value}}
     \includegraphics[width=0.49 \textwidth]{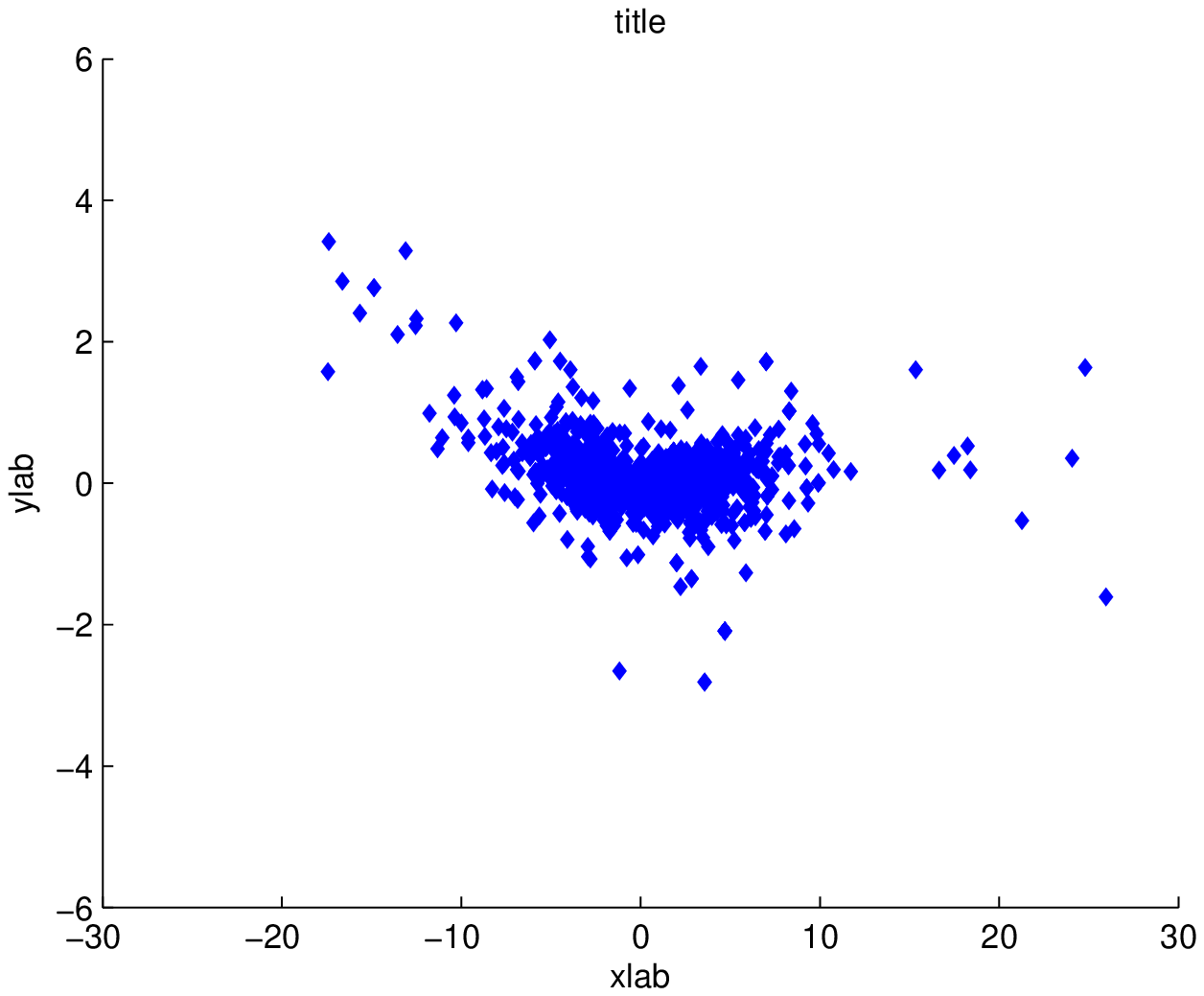}&
     \psfrag{title}[b]{\small{P\&L of Predicting No Abnormal Returns}}
     \psfrag{ylab}[b]{\small{Profit and Loss}}
     \psfrag{xlab}[t]{\small{Change in Stock Value}}
     \includegraphics[width=0.49 \textwidth]{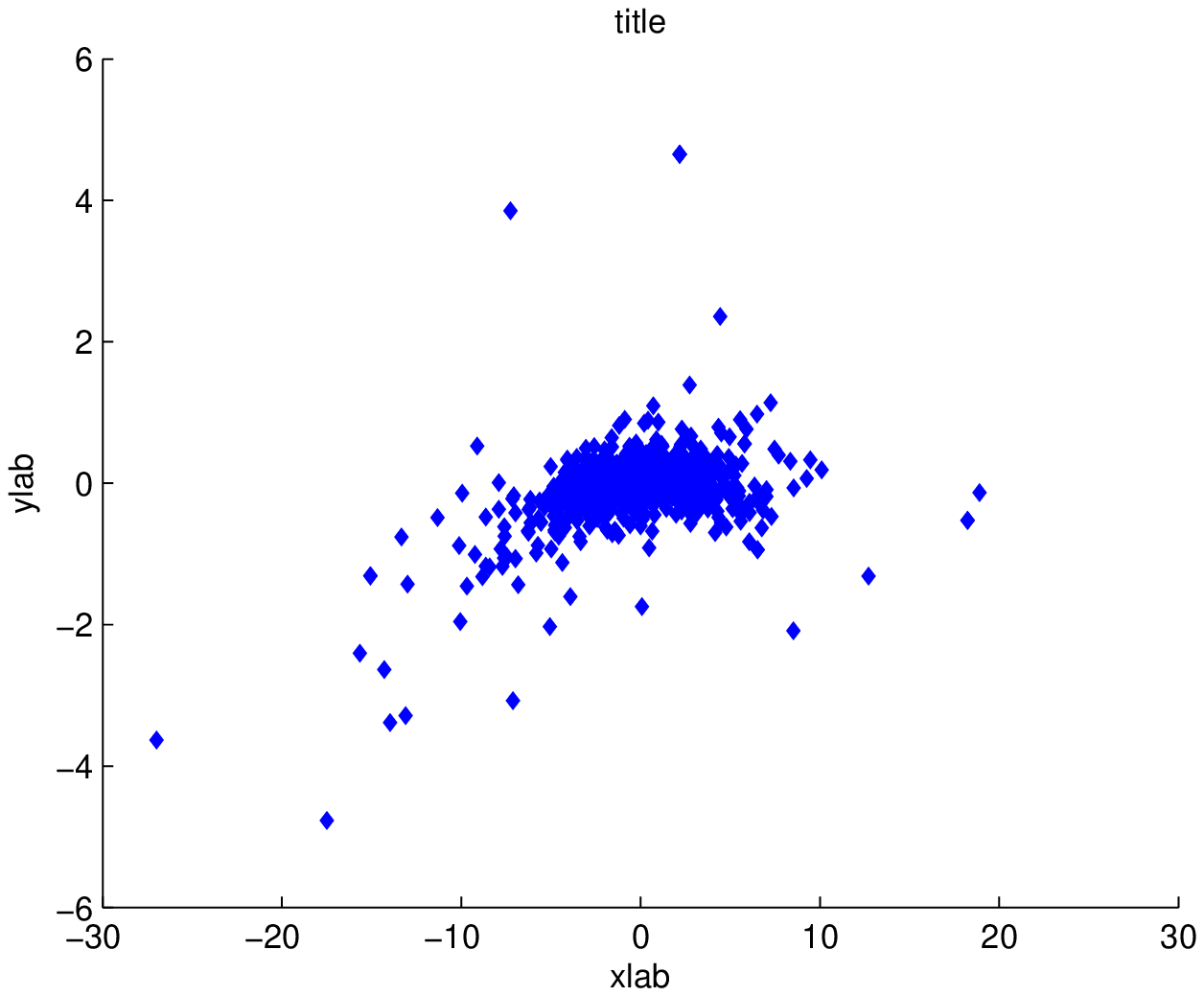}
  \end{tabular}
\caption{Profit and Loss (P\&L) of trading delta-hedged covered call options.  The left figure displays the P\&L of trading on predictions that an abnormal return follows the release of press while the right displays the P\&L resulting from predictions that no abnormal return occurs.  There is a potentially large upside to predicting abnormal returns, however only a limited upside to predicting no movement, while an incorrect prediction of no movement has a potentially large downside.  Text features were used in the related experiments, but figures using returns features do exhibit similar patterns.}
\label{fig:profit_loss_75}
\end{center} \end{figure}

The profit and loss (P\&L) of these strategies is displayed in Figure \ref{fig:profit_loss_75} using the equity and options data.  The left side shows the P\&L of predicting that an abnormal return will occur and the right side shows the P\&L of predicting no price movement.  There is a potentially large upside to predicting abnormal returns, however only a limited upside to predicting no movement, while an incorrect prediction of no movement has a potentially large downside. Text features were used in the related experiments, but figures using returns features do exhibit similar patterns.

\begin{table}[h!]
\begin{center}
\extrarowheight 0.8ex
\begin{tabular}{|c|c|c|c|c|}
\hline
Features & Strategy & Accuracy & Sharpe Ratio & \# Trades \\ \hline
Text & TRADE ALL  &.63  &.75 &3752  \\ \hline
Abs Returns &TRADE ALL  &.54  &-1.01  &3752  \\ \hline
Text &LONG ONLY  &.63 &2.02  &1953  \\ \hline
Abs Returns &LONG ONLY &.54  &1.15  &597  \\ \hline
Text &SHORT ONLY  &.62  & -1.28 & 1670 \\ \hline
Abs Returns &SHORT ONLY  &.54  &-1.95  &3155  \\ \hline
\end{tabular}
\end{center}
\caption{Performance of delta-hedged covered call option strategies.  TRADE ALL makes the appropriate trade based on all predictions, LONG ONLY takes positions only when an abnormal return is predicted, and SHORT ONLY takes positions only when no price movement is predicted.  The $75^{th}$ percentile of absolute returns observed in the training data are used as thresholds for defining abnormal returns.}
\label{table:covered_call_perf}
\end{table}

Table \ref{table:covered_call_perf} displays results for three strategies.  TRADE ALL makes the appropriate trade based on all predictions, LONG ONLY takes positions only when an abnormal return is predicted, and SHORT ONLY takes positions only when no price movement is predicted.  The $75^{th}$ percentile of absolute returns observed in the training data are used as thresholds for defining abnormal returns.  The results imply that the downside of predicting no movement greatly decreases the performance.  The LONG ONLY strategy performs best due to the large upside and only limited downside.  In addition, the number of no movement predictions made using absolute returns features is much larger than when using text.  This is likely the cause of the negative Sharpe ratio for TRADE ALL with absolute returns.  Results using higher thresholds show similar performance trends and the associated P\&L figures have even clearer U-shaped patterns (not displayed).

These results do not account for transaction costs.  Separate experiments set the buy and sell option prices to the highest bid and lowest ask closing prices respectively. Somewhat large spreads mean that the portfolios performed poorly with uniformly negative Sharpe ratios.

\section{Combining text and returns}
\label{sec:multiple_features}
We now discuss multiple kernel learning (MKL), which provides a method for optimally combining text with return data in order to make predictions.  A cutting plane algorithm amenable to large-scale kernels is described and compared with another recent method for MKL.

\subsection{Multiple kernel learning framework}
Multiple kernel learning (MKL) seeks to minimize the upper bound on misclassification probability in (\ref{eq:svm_primal}) by learning an optimal linear combination of kernels (see \citeasnoun{Bous2003}, \citeasnoun{Lanc2004a}, \citeasnoun{Bach2004}, \citeasnoun{Ong2005}, \citeasnoun{Sonn2006}, \citeasnoun{Rako2007}, \citeasnoun{Zien2007}, \citeasnoun{Micc2007}).  The kernel learning problem as formulated in \citeasnoun{Lanc2004a} is written
\BEQ \label{eq:kern_learn_lanc}
\min_{K\in\mathcal{K}} \omega_C(K)
\EEQ
where $\omega_C(K)$ is the minimum of problem (\ref{eq:svm_primal}) and can be viewed as an upper bound on the probability of misclassification.  For general sets $\mathcal{K}$, enforcing Mercer's condition (i.e. $K\succeq 0$) on the kernel $K\in\mathcal{K}$ makes kernel learning a computationally challenging task. The MKL problem in \citeasnoun{Lanc2004a} is a particular instance of kernel learning and solves problem (\ref{eq:kern_learn_lanc}) with
\BEQ \label{eq:MKL_set}
\textstyle \mathcal{K}=\{K\in\symm^n:K=\sum_i{d_iK_i},~\sum_id_i=1,~d\ge0\}
\EEQ
where $K_i\succeq 0$ are predefined kernels. Note that cross-validation over kernel parameters is no longer required because a new kernel is included for each set of desired parameters; however, calibration of the $C$ parameter to SVM is still necessary. The kernel learning problem in (\ref{eq:kern_learn_lanc}) can be written as a semidefinite program when there are no nonnegativity constraints on the kernel weights $d$ in (\ref{eq:MKL_set}) as shown in \citeasnoun{Lanc2004a}.  There are currently no semidefinite programming solvers that can handle large kernel learning problem instances efficiently.  The restriction $d\ge 0$ enforces Mercer's condition and reduces problem (\ref{eq:kern_learn_lanc}) to a quadratically constrained optimization problem
\BEQ  \BA{lll}
&\mbox{maximize} & \alpha^Te-\lambda\\
&\mbox{subject to} & \alpha^Ty=0\\
& & 0\leq\alpha\leq C \\
& & \lambda\ge\frac{1}{2}\alpha^T\mbox{diag}(y)K_i\mbox{diag}(y)\alpha~ \forall i
\EA \label{eq:mkl_quad_constrained} \EEQ
This problem is still numerically challenging for large-scale kernels and several algorithmic approaches have been tested since the initial formulation in \citeasnoun{Lanc2004a}.

The first method, described in \citeasnoun{Bach2004} solves a smooth reformulation of the nondifferentiable dual problem obtained by switching the max and min in problem (\ref{eq:kern_learn_lanc})
\BEQ \label{eq:kern_learn_nondiff} \BA{ll}
\mbox{minimize} & \alpha^Te-\max_i\{
\frac{1}{2}\alpha^T\mbox{diag}(y)K_i\mbox{diag}(y)\alpha\}\\
\mbox{subject to}& \alpha^Ty=0\\
&0\leq\alpha\leq C\\
\EA \EEQ
in the variables $\alpha\in\reals^n$. A regularization term is added in the primal to problem (\ref{eq:kern_learn_nondiff}), which makes the dual a differentiable problem with the same constraints as SVM. A sequential minimal optimization (SMO) algorithm that iteratively optimizes over pairs of variables is used to solve problem~(\ref{eq:kern_learn_nondiff}).

Other approaches for solving larger scale problems are written as a wrapper around an SVM computation. For example, an approach detailed in \citeasnoun{Sonn2006} solves the semi-infinite linear program (SILP) formulation
\BEQ \label{eq:kern_learn_silp} \BA{ll}
\mbox{maximize} & \lambda\\
\mbox{subject to}& \sum_i{d_i}=1\\
& d\ge0\\
&\frac{1}{2}\alpha^T\mbox{diag}(y)(\sum_i{d_iK_i})\mbox{diag}(y)\alpha-\alpha^Te\ge\lambda\\
& \mbox{for all}~\alpha~\mbox{with}~\alpha^Ty=0,0\leq\alpha\leq C\\
\EA \EEQ
in the variables $\lambda\in\reals,d\in\reals^K$.  This problem can be derived from (\ref{eq:kern_learn_lanc}) by moving the objective $\omega_C(K)$ to the constraints.  The algorithm iteratively adds cutting planes to approximate the infinite linear constraints until the solution is found.  Each cut is found by solving an SVM using the current kernel $\sum_i{d_iK_i}$.  This formulation is adapted to multiclass MKL in \citeasnoun{Zien2007} where a similar SILP is solved. The latest formulation in \citeasnoun{Rako2007} is
\BEQ \label{eq:simplemkl}
\min~J(d)~\mbox{s.t.}~\sum_i{d_i}=1,d_i\ge 0\
\EEQ
where
\BEQ
J(d)=\max_{\{0\leq\alpha\leq C,\alpha^Ty=0\}}\alpha^Te-
\frac{1}{2}\alpha^T\mbox{diag}(y)(\sum_i{d_iK_i})\mbox{diag}(y)\alpha
\EEQ
is simply the initial formulation of problem (\ref{eq:kern_learn_lanc}) with the constraints in (\ref{eq:MKL_set}) plugged in.  The authors consider the objective $J(d)$ as a differentiable function of $d$ with gradient calculated as:
\BEQ \label{eq:mkl_grad}
\frac{\partial J}{\partial d_i}=-\frac{1}{2}\alpha^{*T}\mbox{diag}(y)K_i\mbox{diag}(y)\alpha^*
\EEQ
where $\alpha^*$ is the optimal solution to SVM using the kernel $\sum_i{d_iK_i}$.  This becomes a smooth minimization problem subject to box constraints and one linear equality constraint which is solved using a reduced gradient method with a line search.  Each computation of the objective and gradient requires solving an SVM.  Experiments in \citeasnoun{Rako2007} show this method to be more efficient compared to the semi-infinite linear program solved above.  More SVMs are required but warm-starting SVM makes this method somewhat faster.  Still, the reduced gradient method suffers numerically on large kernels as it requires computing many gradients, hence solving many numerically expensive SVM classification problems.

\subsection{Multiple kernel learning via an analytic center cutting plane method}
We next detail a more efficient algorithm for solving problem (\ref{eq:simplemkl}) that requires far less SVM computations than gradient descent methods.  The analytic center cutting plane method (ACCPM) iteratively reduces the volume of a localizing set $\mathcal{L}$ containing the optimum using \emph{cuts} derived from a first order convexity property until the volume of the reduced localizing set converges to the target precision. At each iteration $i$, a new center is computed in a smaller localizing set $\mathcal{L}_i$ and a cut through this point is added to split $\mathcal{L}_i$ and create $\mathcal{L}_{i+1}$.  The method can be modified according to how the center is selected; in our case the center selected is the analytic center of $\mathcal{L}_i$ defined below. Note that this method does not require differentiability but still exhibits linear convergence.

We set $\mathcal{L}_0=\{d\in\reals^n | \sum_i{d_i}=1,d_i\ge 0\}$ which we can write as $\{d\in\reals^n | A_0d\leq b_0\}$ (the single equality constraint can be removed by a different parameterization of the problem) to be our first localization set for the optimal solution.  Our method is then described as Algorithm~\ref{alg:accpm_2} below (see \citeasnoun{Bert1999} for a more complete reference on cutting plane methods). The complexity of each iteration breaks down as follows.
\BIT
\item \emph{Step 1.} This step computes the analytic center of a polyhedron and can be solved in $O(n^3)$ operations using interior point methods for example.
\item \emph{Step 2.} This step updates the polyhedral description.  Computation of $\nabla J(d)$ requires a single SVM computation which can be speeded up by warm-starting with the SVM solution of the previous iteration.

\item \emph{Step 3.} This step requires ordering the constraints according to their relevance in the localization set.  One relevance measure for the $j^{th}$ constraint at iteration $i$ is
\BEQ
\frac{a_j^T\nabla^2f(d_i)^{-1}a_j}{(a_j^td_i-b_j)^2}
\EEQ
where $f$ is the objective function of the analytic center problem. Computing the hessian is easy: it requires matrix multiplication of the form $A^TDA$ where $A$ is $m\times n$ (matrix multiplication is kept inexpensive in this step by pruning redundant constraints) and $D$ is diagonal.

\item \emph{Step 4.}  An explicit duality gap can be calculated at no extra cost at each iteration because we can obtain the dual MKL solution without further computations.  The duality gap (as shown in \citeasnoun{Rako2007}) is:
\BEQ \label{eq:mkl_dualitygap}
\max_i{(\alpha^{*T}\mbox{diag}(y)K_i\mbox{diag}(y)\alpha^*)}-\alpha^{*T}\mbox{diag}(y)(\sum_id_iK_i)\mbox{diag}(y)\alpha^*
\EEQ
where $\alpha^*$ is the optimal solution to SVM using the kernel $\sum_i{d_iK_i}$.
\EIT

\begin{algorithm}[h]
\caption{Analytic center cutting plane method}
\begin{algorithmic} [1]
\STATE Compute $d_i$ as the analytic center of $\mathcal{L}_i$=$\{d\in\reals^n | A_id\leq b_i\}$ by solving:
    \[
    d_{i+1}=\argmin_{y\in\reals^n} ~ -\sum_{i=1}^m \mbox{log}(b_i-a_i^Ty)
    \]
    where $a_i^T$ represents the $i^{th}$ row of coefficients from $A_i$ in $\mathcal{L}_i$, $m$ is the number of rows in $A_i$, and $n$ is the dimension of $d$ (the number of kernels).
\STATE Compute $\nabla J(d)$ from (\ref{eq:mkl_grad}) at the center $d_{i+1}$ and update the (polyhedral) localization set:
    \[
    \mathcal{L}_{i+1}=\mathcal{L}_i \cap \{d\in\reals^n | \nabla J(d_{i+1})(d-d_{i+1}) \ge 0\}
    \]
\STATE If $m\ge 3n$, reduce the number of constraints to $3n$.
\STATE If $\mbox{gap} \le \epsilon$ stop, otherwise go back to step 1.
\end{algorithmic}
\label{alg:accpm_2}
\end{algorithm}

\begin{figure} [h!]
\begin{tabular*}{\textwidth}{@{\extracolsep{\fill}}cc}
\psfrag{gap}[b][t]{\small{Duality Gap}}
\psfrag{iter}[t][b]{\small{Iteration}}
\includegraphics[width=.50 \textwidth]{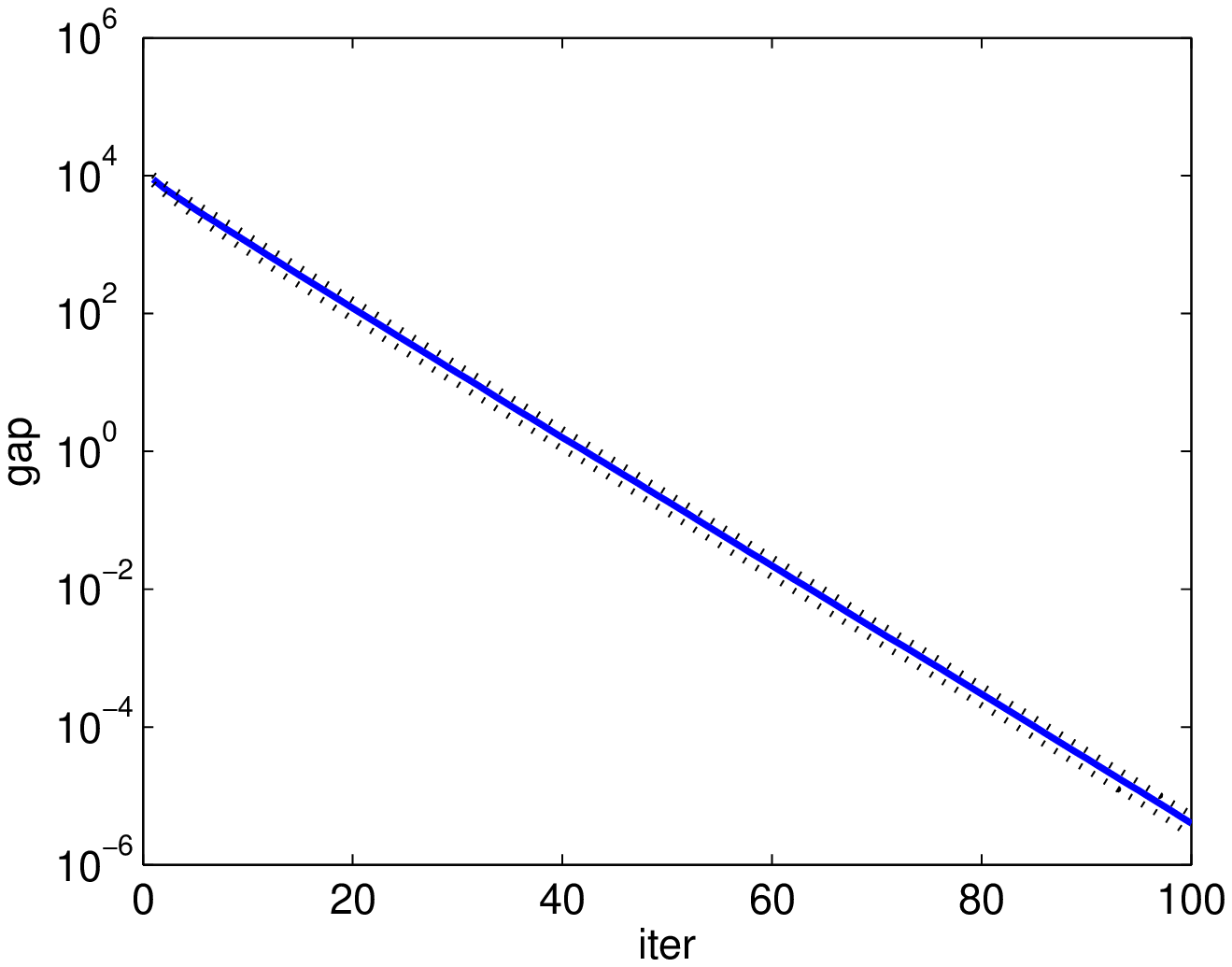} &
\psfrag{time}[b][t]{\small{Time (seconds)}} \psfrag{kern}[t][b]{\small{Number of Kernels}}
\includegraphics[width=.50 \textwidth]{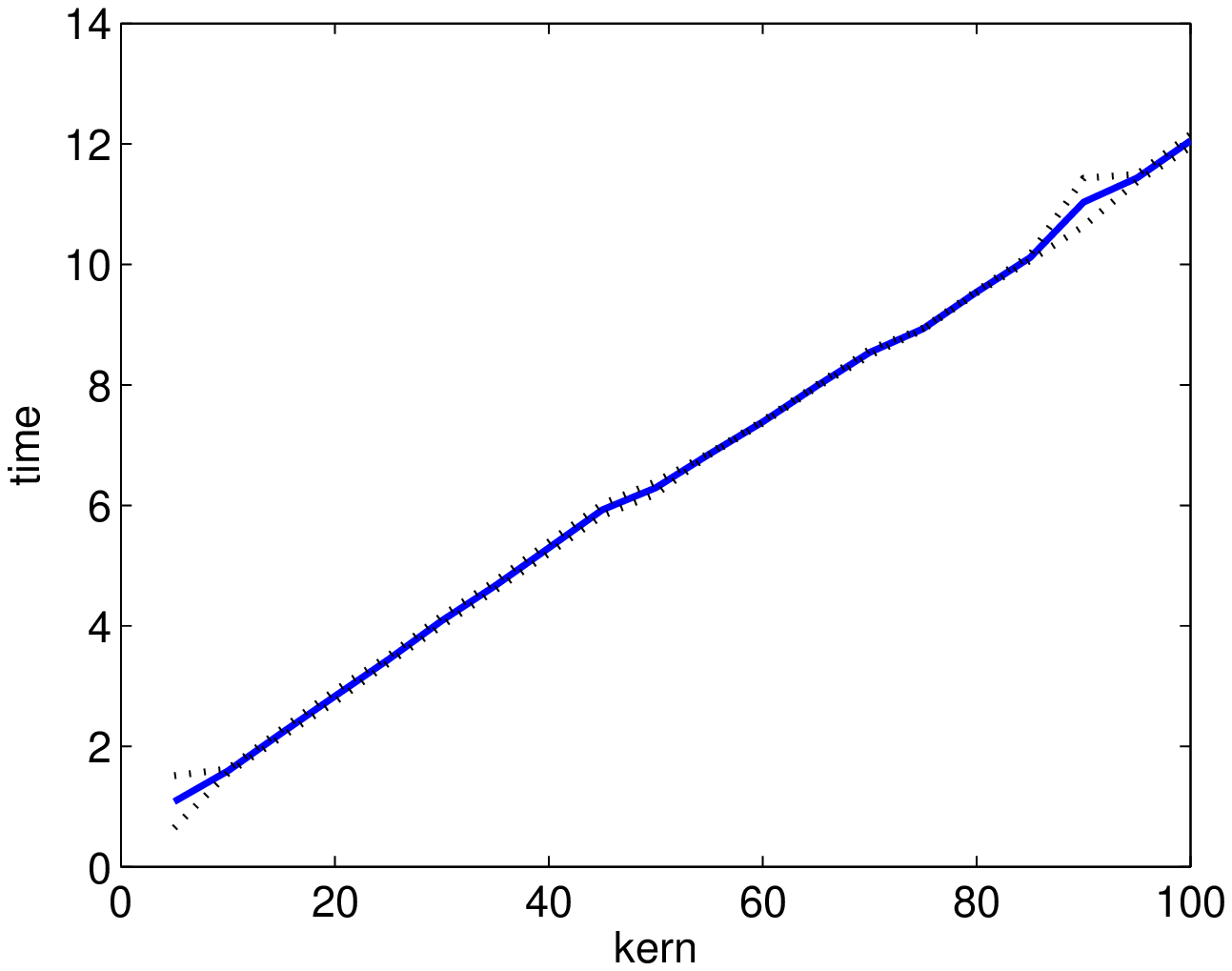}
\end{tabular*}
\caption{The convergence semilog plot for ACCPM (left) shows average gap versus iteration number. We plot CPU time  for the first 10 iterations versus number of kernels (right).  Both plots give averages over 20 experiments with dashed lines at plus and minus one standard deviation.  In all these experiments, ACCPM converges linearly to a high precision.}
\label{fig:accpmConvergence}
\end{figure}

\paragraph{Complexity.}  ACCPM is provably convergent in $O(n(\log 1/\epsilon)^2)$ iterations when using a cut elimination scheme as in \citeasnoun{Atki1995} which keeps the complexity of the localization set bounded. Other schemes are available with slightly different complexities: $O(n^2/\epsilon^2)$ is achieved in \citeasnoun{Goff2002} using (cheaper) approximate centers for example.  In practice, ACCPM usually converges linearly as seen in Figure \ref{fig:accpmConvergence} (left) which uses kernels of dimension 500 on text data.  To illustrate the affect of increasing the number of kernels on the analytic center problem, Figure \ref{fig:accpmConvergence} (right) shows CPU time increasing as the number of kernels increases.

Gradient methods such as the reduced gradient method used in simpleMKL converge linearly (see \citeasnoun{Luen2003}), but require expensive line searches.  Therefore, while gradient methods may sometimes converge linearly at a faster rate than ACCPM on certain problems, they are often much slower due to the need to solve many SVM problems per iteration. Empirically, gradient methods tend to require many more gradient evaluations than the localization techniques discussed here. ACCPM computes the objective and gradient exactly once per iteration and the analytic center problem remains relatively cheap with respect to the SVM computation because the dimension of the analytic centering problem (i.e. the number of kernels) is small in our application. Thresholding small kernel weights in MKL to zero can further reduce the dimension of the analytic center problem.

\subsection{Computational Savings}
As described above, ACCPM computes one SVM computation per iteration and converges linearly.  We compare this method, which we denote accpmMKL, with the simpleMKL algorithm which uses a reduced gradient method and also converges fast but computes many more SVMs to perform line searches. The SVMs in the line search are speeded up using warm-starting as described in \citeasnoun{Rako2007} but in practice, we observe that savings in MKL from warm-starting often do not suffice to make this gradient method more efficient than ACCPM.

Few kernels are usually required in MKL because most kernels can be eliminated more efficiently beforehand using cross-validation, hence we use several families of kernels (linear, gaussian, and polynomial) but very few kernels from each family.  Each experiment uses one linear kernel and the same number of gaussian and polynomial kernels giving a total of 3, 7, and 11 kernels (each normalized to unit trace) in each experiment.  We set the duality gap to .01 (a very loose gap) and $C$ to 1000 (after cross-validation for $C$ ranging between 500 and 5000) for each experiment in order to compare the algorithms on identical problems.  For fairness, we compare simpleMKL with our implementation of accpmMKL using the same SVM package in simpleMKL  which allows warm-starting (The SVM package in simpleMKL is based on the SVM-KM toolbox \cite{SVM-KMToolbox} and implemented in Matlab.).  In the final column, we also give the running time for accpmMKL using the LIBSVM solver without warm-starting.  The following tables demonstrate computational efficiency and do not show predictive performance; both algorithms solve the same optimization problem with the same stopping criterion. High precision for MKL does not significantly increase prediction performance.  Results are averages over 20 experiments done on Linux 64-bit servers with 2.6 GHz CPUs.

Table \ref{table:text_time_stats} shows that ACCPM is more efficient for the multiple kernel learning problem in a text classification example.  Savings from warm-starting SVM in simpleMKL do not overcome the benefit of fewer SVM computations at each iteration in accpmMKL.  Furthermore, using a faster SVM solver such as LIBSVM produces better performance even without warm-starting.  The number of kernels used in accpmMKL is higher than with simpleMKL because of the very loose duality gap here.  The reduced gradient method of simpleMKL often stops at a much higher precision because the gap is checked after a line search that can achieve high precision in a single iteration and it is this higher precision that reduces the number of kernels. However, for a slightly higher precision, simpleMKL will often stall or converge very slowly; the method is very sensitive to the target precision. The accpmMKL method stops at the desired duality (meaning more kernels) because the gap is checked at each iteration during the linear convergence; however, the convergence is much more stable and consistent for all data sets.  For accpmMKL, the number of SVMs is equivalent to the number of iterations.

\begin{table}[h!]
\begin{center}
\small{
\begin{tabular}{|c|c|c|c|c|c|c|c|c|c|}
\hline
  & Max& \multicolumn{4}{|c|}{simpleMKL} & \multicolumn{4}{|c|}{accpmMKL}\\ \hline
   Dim&\# Kern&\# Kern&\# Iters&\# SVMs&Time&\# Kern&\# SVMs&Time&Time (LIBSVM)\\
\hline
\multirow{3}{*}{500}
&3&2.0&3.4&27.2&48.6&3.0&7.1&\textbf{13.7}&\textbf{0.6}\\
&7&2.6&3.4&39.5&47.9&7.0&12.0&\textbf{15.5}&\textbf{1.8}\\
&11&3.6&3.2&41.0&37.3&10.9&15.3&\textbf{17.4}&\textbf{3.3}\\
\hline
\multirow{3}{*}{1000}
&3&2.0&2.0&29.3&164.5&3.0&6.3&\textbf{36.7}&\textbf{2.4}\\
&7&2.4&3.6&53.3&240.3&6.8&11.7&\textbf{40.0}&\textbf{6.8}\\
&11&3.9&3.6&57.8&214.6&10.6&14.9&\textbf{48.1}&\textbf{12.7}\\
\hline
\multirow{3}{*}{2000}
&3&2.0&1.0&24.0&265.8&3.0&5.0&\textbf{79.4}&\textbf{7.2}\\
&7&3.3&1.5&30.4&209.6&7.0&10.5&\textbf{110.5}&\textbf{25.2}\\
&11&6.0&2.3&40.5&253.2&11.0&14.4&\textbf{141.4}&\textbf{46.5}\\
\hline
\multirow{3}{*}{3000}
&3&2.0&1.0&24.0&435.5&3.0&6.0&\textbf{248.9}&\textbf{17.9}\\
&7&4.0&2.0&38.0&591.4&7.0&6.8&\textbf{221.7}&\textbf{39.0}\\
&11&6.0&2.0&39.8&648.9&11.0&8.0&\textbf{244.8}&\textbf{66.8}\\
\hline
\end{tabular}
}
\end{center}
\caption{Numerical performance of simpleMKL versus accpmMKL for classification on text classification data.  accpmMKL outperforms simpleMKL in terms of SVM iterations and time. Using LIBSVM to solve SVM problems further enhances performance.  Results are averages over 20 runs.  Experiments are done using the SVM solver in the simpleMKL toolbox except for the final column which uses LIBSVM.  Time is in seconds.  Dim is the number of training samples in each kernel.}
\label{table:text_time_stats}
\end{table}

Table \ref{table:mushroom_time_stats} shows an example where accpmMKL is outperformed by simpleMKL.  This occurs when the classification task is extremely easy and the optimal mix of kernels is a singleton.  In this case, simpleMKL converges with fewer SVMs.  Note though that accpmMKL with LIBSVM is still faster here.  Both examples illustrate that simpleMKL trains many more SVMs whenever the optimal mix of kernels includes more than one input kernel.  Overall, accpmMKL has the advantages of consistent convergence rates for all data sets, fewer SVM computations for relevant data sets, and the ability to achieve high precision targets.

\begin{table}[h!]
\begin{center}
\small{
\begin{tabular}{|c|c|c|c|c|c|c|c|c|c|}
\hline
  & Max& \multicolumn{4}{|c|}{simpleMKL} & \multicolumn{4}{|c|}{accpmMKL}\\ \hline
   Dim&\# Kern&\# Kern&\# Iters&\# SVMs&Time&\# Kern&\# SVMs&Time&Time (LIBSVM)\\
\hline
\multirow{3}{*}{500}
&3&2.0&1.9&32.8&22.3&2.0&11.1&\textbf{5.8}&\textbf{0.8}\\
&7&1.6&2.8&22.6&19.2&7.0&14.7&\textbf{3.7}&\textbf{1.9}\\
&11&1.0&2.0&11.6&\textbf{7.1}&8.2&20.4&9.1&\textbf{4.1}\\
\hline
\multirow{3}{*}{1000}
&3&2.0&2.0&32.6&70.6&3.0&5.0&\textbf{8.7}&\textbf{1.5}\\
&7&1.0&2.0&9.9&\textbf{10.6}&7.0&15.7&17.2&\textbf{8.2}\\
&11&1.0&2.0&11.6&\textbf{38.4}&8.0&21.0&48.6&\textbf{16.8}\\
\hline
\multirow{3}{*}{2000}
&3&1.0&1.0&4.0&\textbf{36.5}&3.0&6.0&41.8&\textbf{7.0}\\
&7&1.0&2.0&10.3&\textbf{54.0}&7.0&16.0&85.5&\textbf{34.0}\\
&11&1.0&2.0&12.1&\textbf{261.7}&8.0&21.0&294.8&\textbf{67.5}\\
\hline
\multirow{3}{*}{3000}
&3&1.0&1.0&4.0&\textbf{89.4}&3.0&6.0&100.9&\textbf{15.1}\\
&7&1.0&2.0&10.5&\textbf{158.3}&7.0&16.0&235.4&\textbf{79.9}\\
&11&1.0&2.0&12.2&\textbf{925.9}&8.0&21.0&959.5&\textbf{163.4}\\
\hline
\end{tabular}
}
\end{center}
\caption{Numerical performance of simpleMKL versus accpmMKL for classification on \textbf{UCI Mushroom Data}.  simpleMKL outperforms accpmMKL when the classification task is very easy, demonstrated by optimality of a single kernel, but otherwise performs slower.  Experiments are done using the SVM solver in the simpleMKL toolbox except for the final column which uses LIBSVM.  Time is in seconds.  Dim is the number of training instances in each kernel.}
\label{table:mushroom_time_stats}
\end{table}

\subsection{Predicting abnormal returns with text \emph{and} returns}
Multiple kernel learning is used here to combine text with returns data in order to predict abnormal equity returns.  Kernels $K_1,...,K_i$ are created using only text features as done in Section \ref{subsec:pred_with_svm} and additional kernels $K_{i+1},...,K_d$ are created from a time series of absolute returns.  Experiments here use one linear and four Gaussian kernels, each normalized to have unit trace, for each feature type.  The MKL problem is solved using $K_1,...K_d$, two linear kernels based on time of day and day of week, and an additional identity matrix in $\mathcal{K}$ described by (\ref{eq:MKL_set}); hence we obtain a single optimal kernel $K^*=\sum_i{d_i^*K_i}$ that is a convex combination of the input kernels.  The same technique (referred to as data fusion) was applied in \citeasnoun{Lanc2004b} to combine protein sequences with gene expression data in order to recognize different protein classes.

\begin{figure}[h!] \begin{center}
  \begin{tabular} {cc}
     \psfrag{title}[b]{\small{Accuracy using Multiple Kernels}}
     \psfrag{acc}[b]{\small{Accuracy}}
     \psfrag{min}[t]{\small{Minutes}}
     \includegraphics[width=0.49 \textwidth]{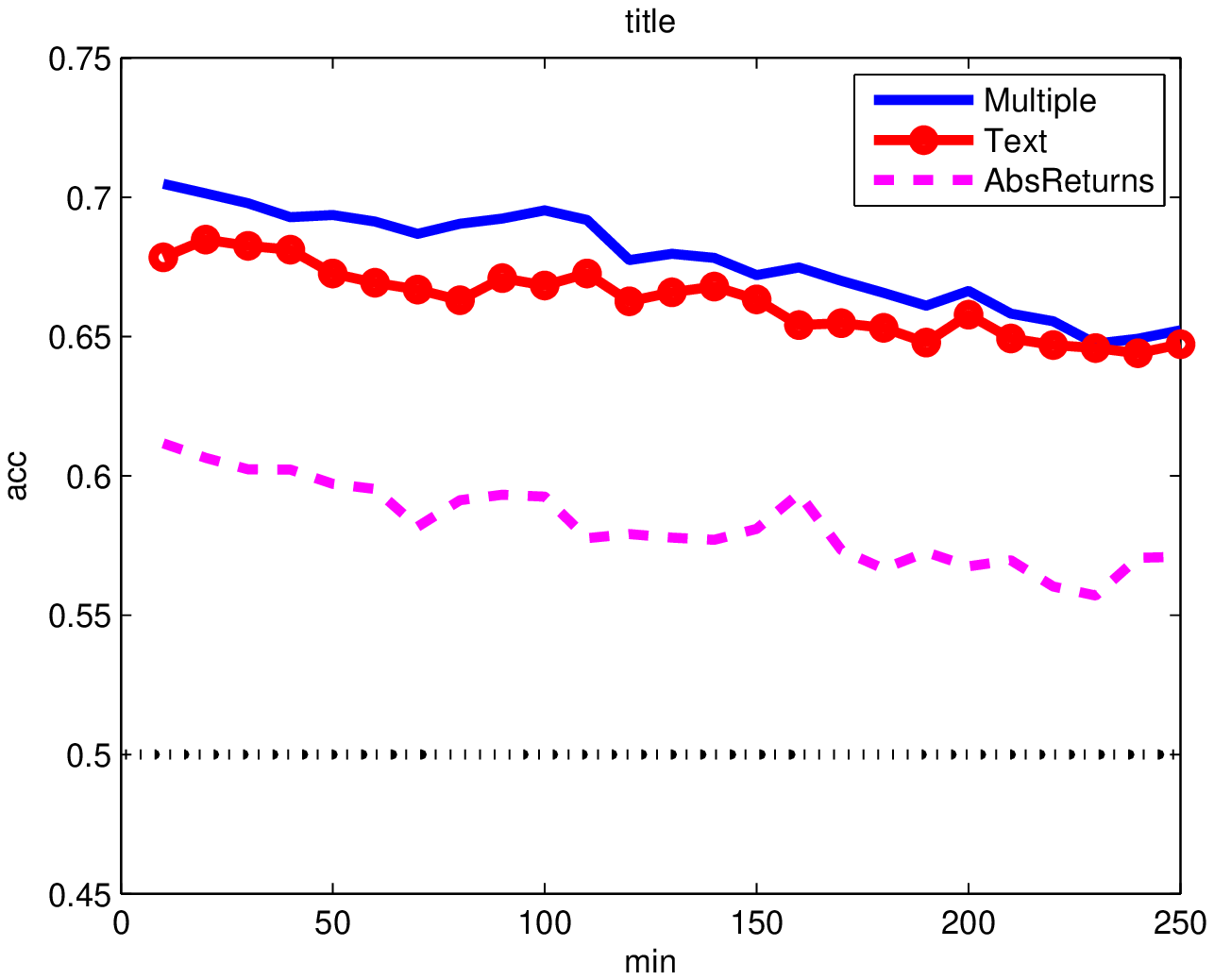}&
    \psfrag{title}[b]{\small{Sharpe Ratio using Multiple Kernels}}
    \psfrag{sharpe}[b]{\small{Sharpe Ratio}}
    \psfrag{min}[t]{\small{Minutes}}
    \includegraphics[width=0.49 \textwidth]{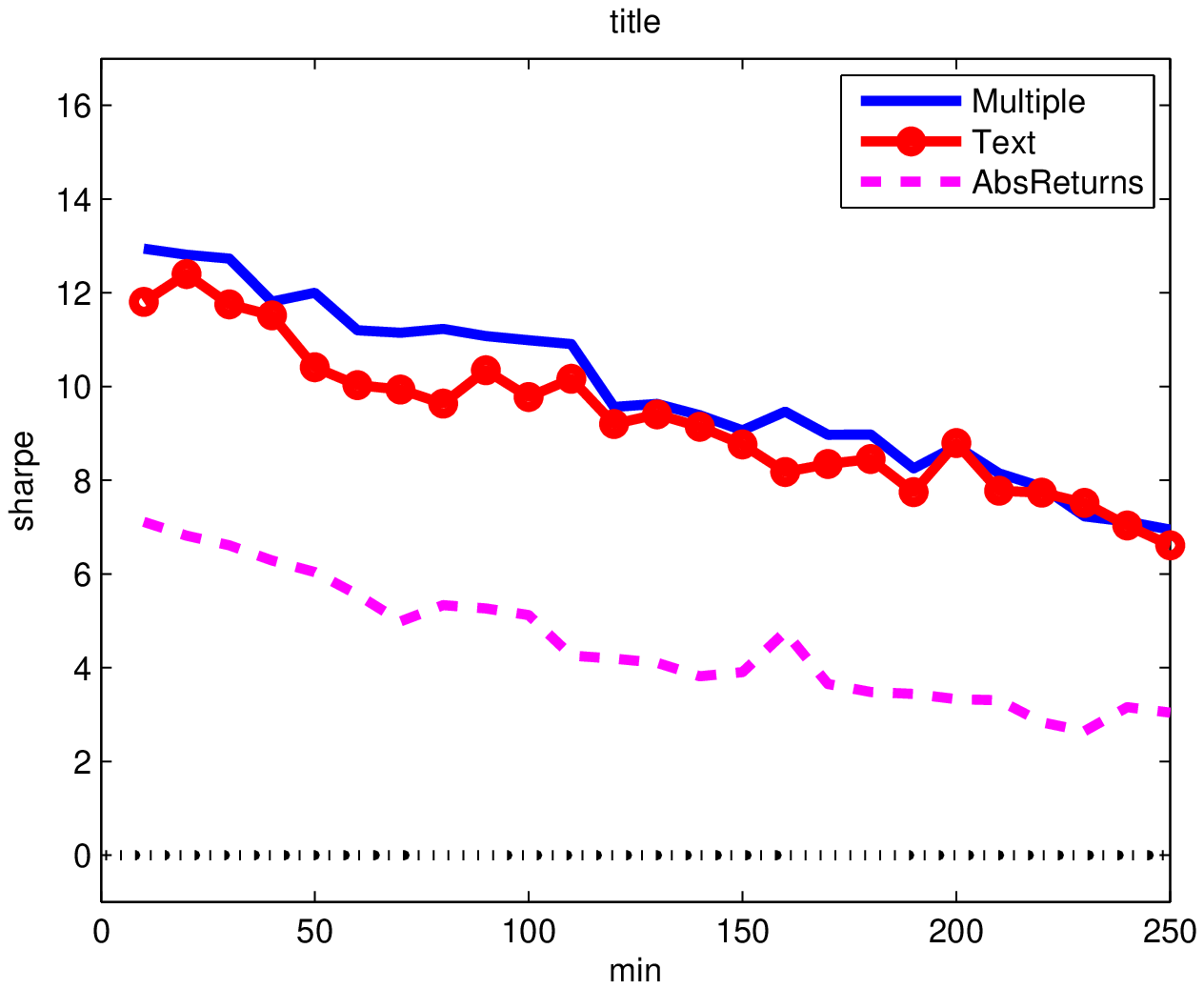}
  \end{tabular}
\caption{Accuracy and sharpe ratio using multiple kernels.  MKL mixes 13 possible kernels (1 linear text, 1 linear absolute returns, 4 gaussian text, 4 gaussian absolute returns, 1 linear time of day, 1 linear day of week, 1 identity matrix).  Each point \emph{z} on the x-axis corresponds to predicting an abnormal return \emph{z} minutes after each press release is issued.  The $75^{th}$ percentile of absolute returns observed in the training data is used as the threshold for defining an abnormal return.}
\label{fig:multiple_kernels_move}
\end{center} \end{figure}

Performance using the $75^{th}$ percentile of absolute returns as a threshold for abnormality are displayed in Figure \ref{fig:multiple_kernels_move}.  Results from Section \ref{subsec:pred_with_svm} that use SVM with a text and absolute returns linear kernels are superimposed with the performance when combining text, absolute returns, and time stamps.  While predictions using only text or returns exhibit good performance, combining them significantly improves performance in both accuracy and annualized daily Sharpe ratio.

\begin{figure}[h!] \begin{center}
     \psfrag{13kerns}[b]{\small{Coefficients with Multiple Kernels $75^{th}$ \%}}
     \psfrag{coeff}[b]{\small{Coefficients}}
     \psfrag{min}[t]{\small{Minutes}}
     \includegraphics[width=0.49 \textwidth]{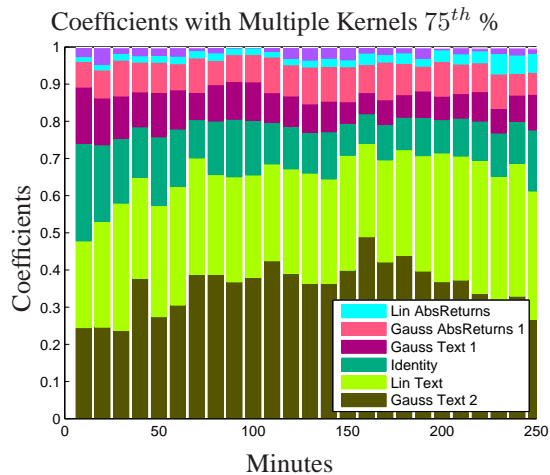}
\caption{Optimal kernel coefficients when using when using 13 possible kernels (1 linear text, 1 linear absolute returns, 4 gaussian text, 4 gaussian absolute returns, 1 linear time of day, 1 linear day of week, 1 identity matrix) with $75^{th}$ percentile threshold to define abnormal returns.  Only the top 5 kernels are labeled.  Each point \emph{z} on the x-axis corresponds to predicting an abnormal return \emph{z} minutes after each press release is issued.}
\label{fig:13kernels_coeff_move}
\end{center} \end{figure}

We next analyze the impact of the various kernels.  Figure \ref{fig:13kernels_coeff_move} displays the optimal kernel weights $d_i$ found from solving (\ref{eq:simplemkl}) at each time horizon (weights are averaged from results over each window).  Kernel weights are represented as colored fractions of a single bar of length one.  The five kernels with the largest coefficients are two gaussian text kernels, a linear text kernel, the identity kernel, and one gaussian absolute returns kernels. Note that the magnitudes of the coefficients are not perfectly indicative of importance of the respective features.  Hence, the optimal mix of kernels here supports the above evidence that mixing news with absolute returns improves performance.  Another important observation is that kernel weights remain relatively constant over time.  Each bar of kernel weights corresponds to an independent classification task (i.e. each predicts abnormal returns at different times in the future) and the persistent kernel weights imply that combining important kernels detects a meaningful signal beyond that found by using only text or return features.

\begin{figure}[h!] \begin{center}
  \begin{tabular} {cc}
     \psfrag{title}[b]{\small{Accuracy using Multiple Kernels}}
     \psfrag{acc}[b]{\small{Accuracy}}
     \psfrag{min}[t]{\small{Minutes}}
     \includegraphics[width=0.49 \textwidth]{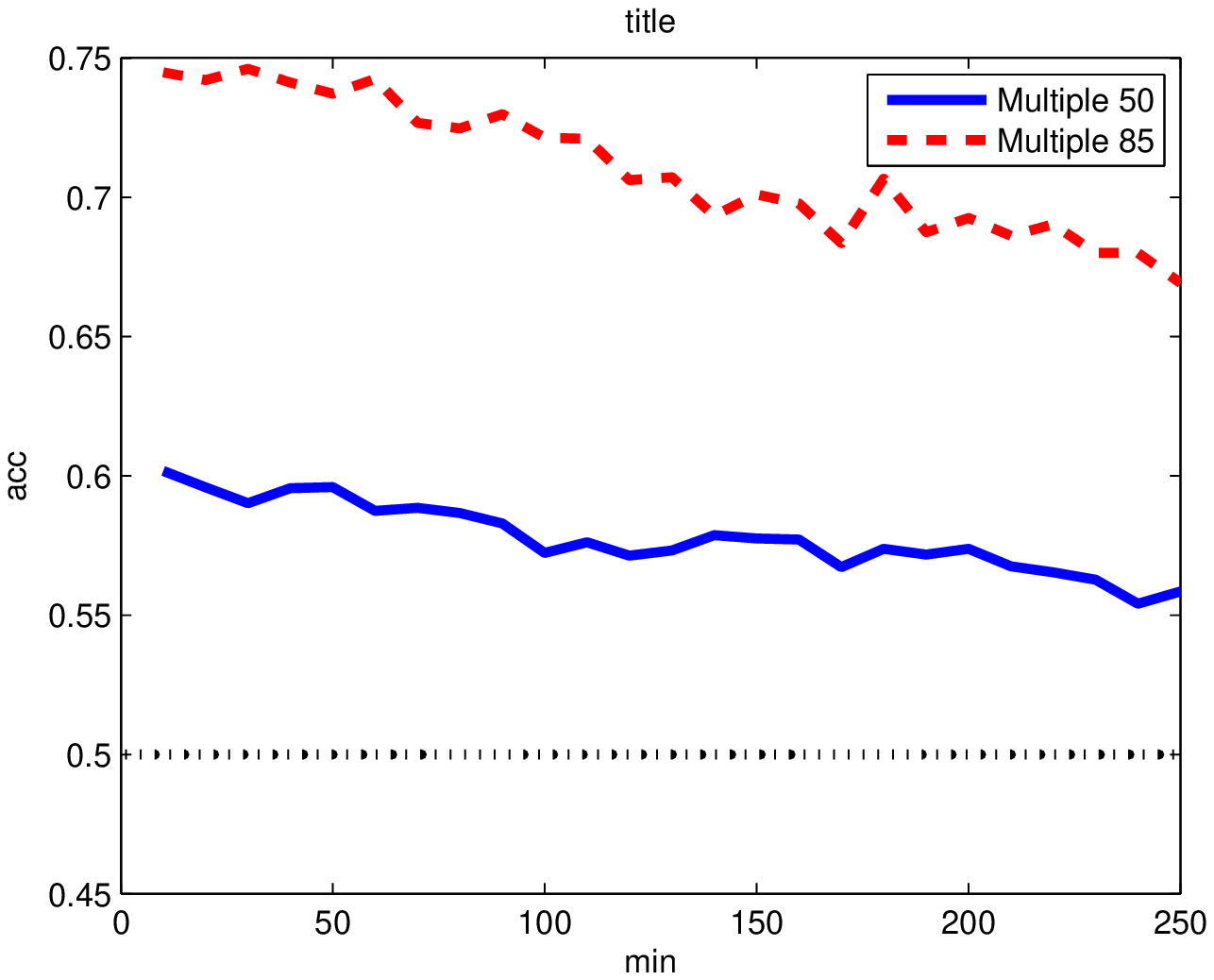}&
     \psfrag{title}[b]{\small{Sharpe Ratio using Multiple Kernels}}
     \psfrag{sharpe}[b]{\small{Sharpe Ratio}}
     \psfrag{min}[t]{\small{Minutes}}
     \includegraphics[width=0.49 \textwidth]{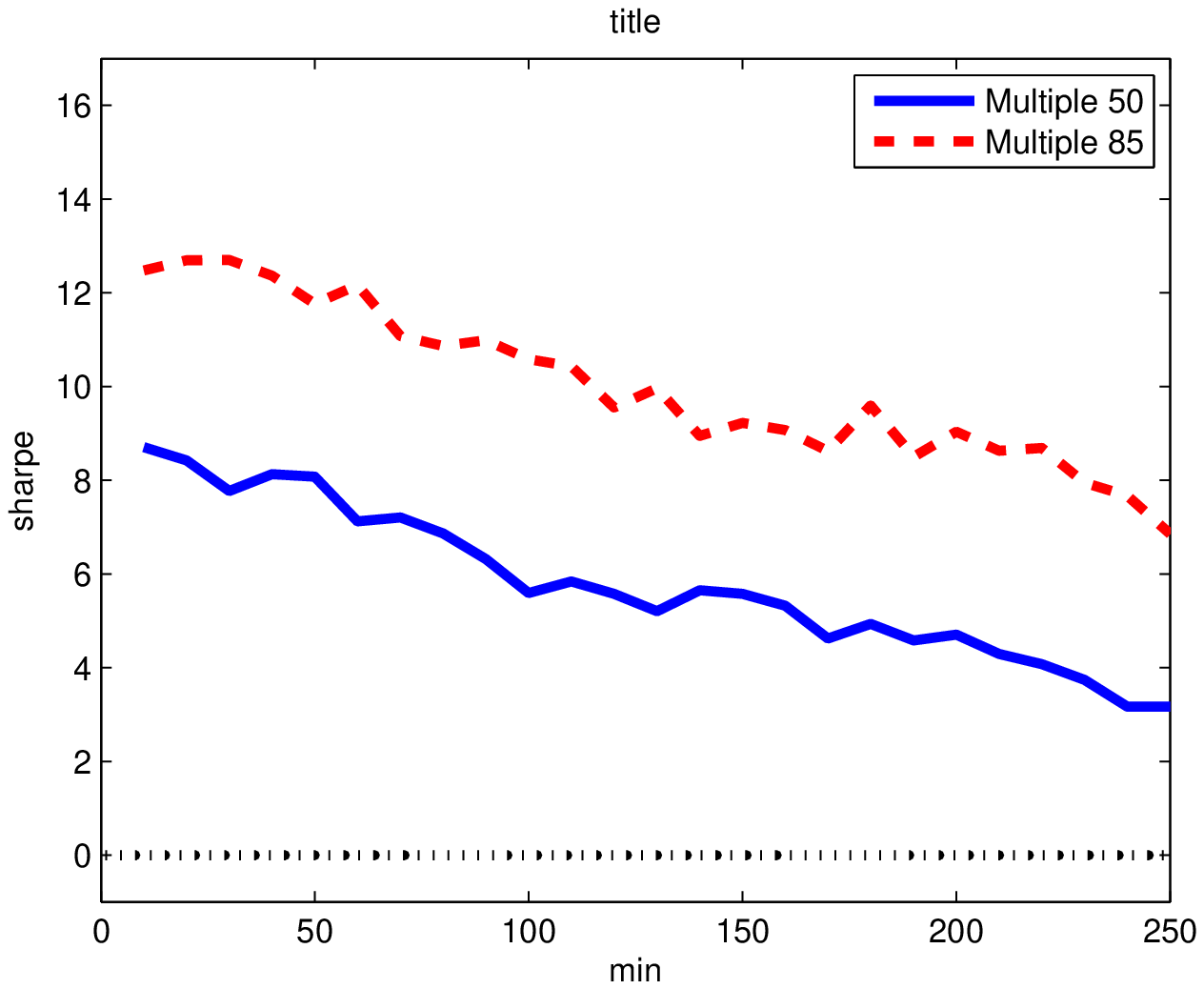}
  \end{tabular}
\caption{Accuracy and annualized daily sharpe ratio for predicting abnormal returns using multiple kernels.  Each point \emph{z} on the x-axis corresponds to predicting an abnormal return \emph{z} minutes after each press release is issued.  The $50^{th}$ and $85^{th}$ percentiles of absolute returns are used to define abnormal returns.}
\label{fig:multiple_kernels_movement_50_85}
\end{center} \end{figure}

\begin{figure}[h!] \begin{center}
  \begin{tabular} {cc}
     \psfrag{13kerns}[b]{\small{Coefficients with Multiple Kernels $50^{th}$ \%}}
     \psfrag{coeff}[b]{\small{Coefficients}}
     \psfrag{min}[t]{\small{Minutes}}
     \includegraphics[width=0.49 \textwidth]{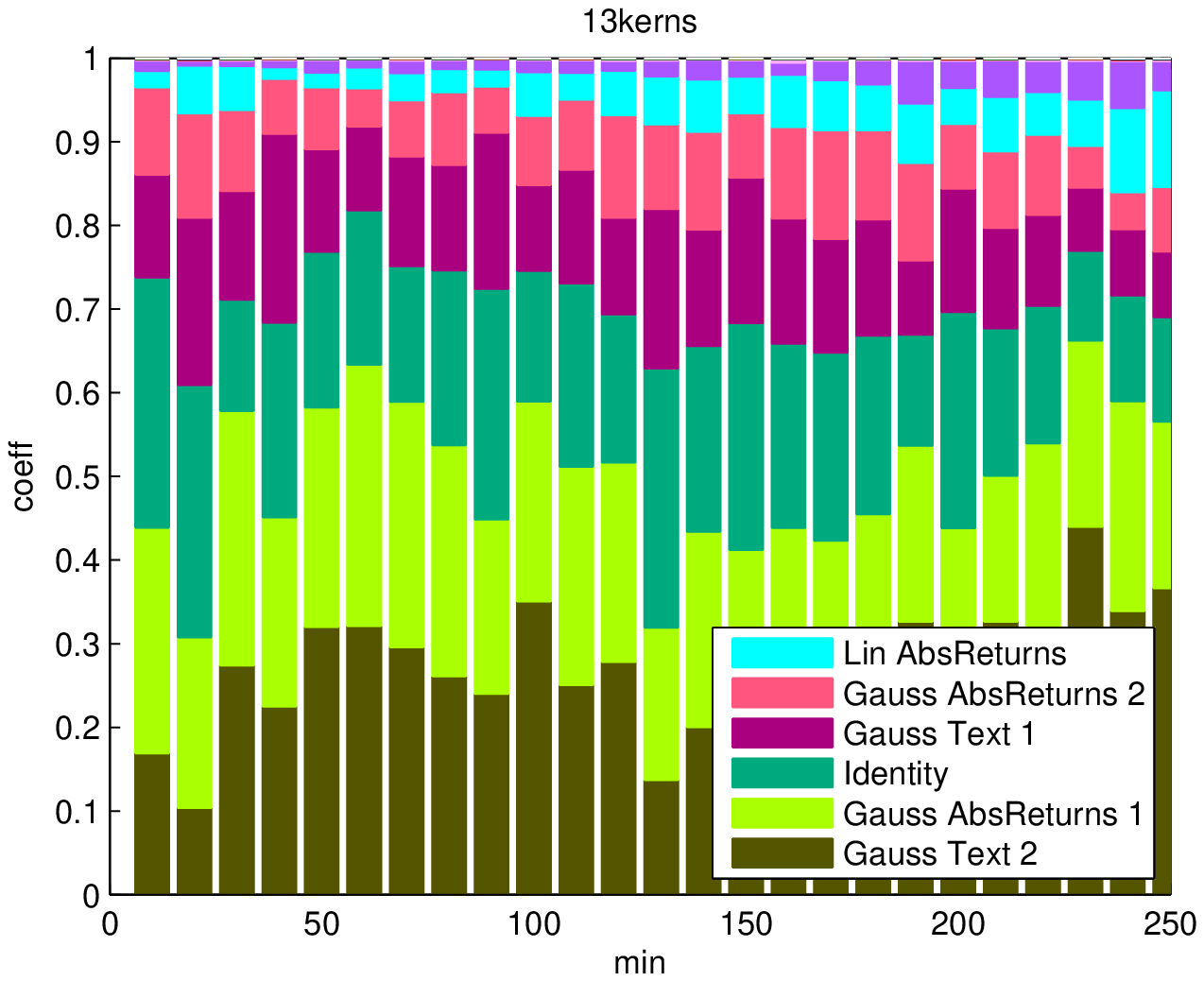}&
     \psfrag{13kerns}[b]{\small{Coefficients with Multiple Kernels $85^{th}$ \%}}
     \psfrag{coeff}[b]{\small{Coefficients}}
     \psfrag{min}[t]{\small{Minutes}}
     \includegraphics[width=0.49 \textwidth]{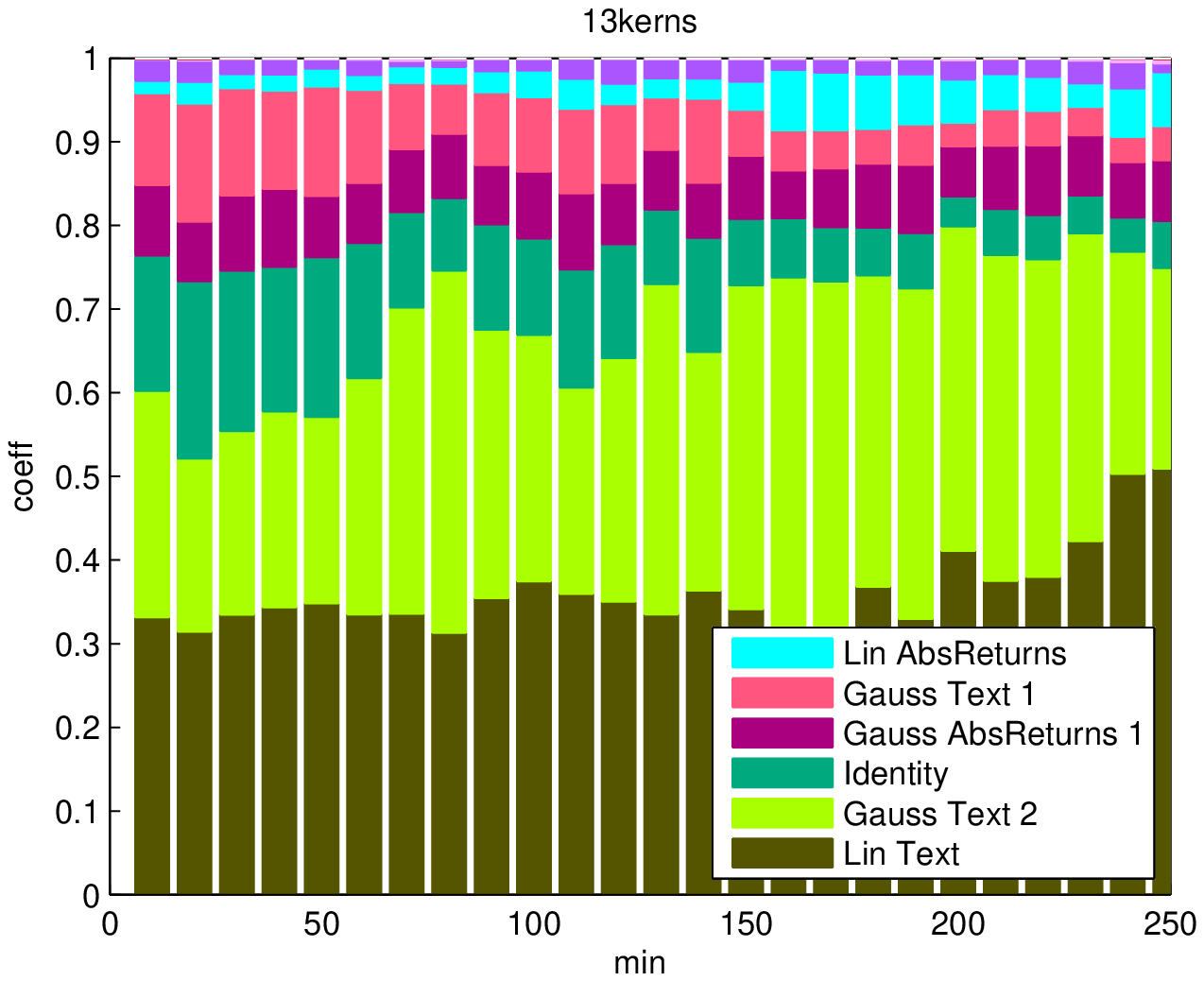}
  \end{tabular}
\caption{Optimal kernel coefficients when using 13 possible kernels (1 linear text, 1 linear absolute returns, 4 gaussian text, 4 gaussian absolute returns, 1 linear time of day, 1 linear day of week, 1 identity matrix) with $50^{th}$ and $85^{th}$ percentiles as thresholds.  Only the top 5 kernels are labeled.  Each point \emph{z} on the x-axis corresponds to predicting an abnormal return \emph{z} minutes after each press release is issued.}
\label{fig:10kernels_coeff_move50_85}
\end{center} \end{figure}

Figure \ref{fig:multiple_kernels_movement_50_85} shows the performance of using multiple kernels for predicting abnormal returns when we change the threshold to the $50^{th}$ and $85^{th}$ percentiles of absolute returns in the training data.  In both cases, there is a slight improvement in performance from using single kernels.  Figure \ref{fig:10kernels_coeff_move50_85} displays the optimal kernel weights for these experiments, and, indeed, both experiments use a mix of text and absolute returns.  Previously, text was shown to have more predictability with a higher threshold while absolute returns performed better with a lower threshold.  Kernel weights here versus those with the $75^{th}$ percentile threshold reflect this observation.

\subsection{Sensitivity of MKL}
Successful performance using multiple kernel learning is highly dependent on a proper choice of input kernels.  Here, we show that high accuracy of the optimal mix of kernels is not crucial for good performance, while including the optimal kernels in the mix is necessary.  In addition, we show that MKL is insensitive to the inclusion of kernels with no information (such as random kernels).  The following four experiments with different kernels sets exemplify these observations.  First, only linear kernels using text, absolute returns, time of day, and day of week are included.  Next, an equal weighting ($d_i=1/13$) for thirteen kernels (one linear and four gaussian each from text and absolute returns, one linear for each time of day and day of week, and an identity kernel) is used.  Another test performs MKL using the same thirteen kernels in addition to three random kernels and a final experiment uses four bad gaussian kernels (two text and two absolute returns).

\begin{figure}[h!] \begin{center}
  \begin{tabular} {cc}
     \psfrag{title}[b]{\small{Accuracy for Various Tests}}
     \psfrag{acc}[b]{\small{Accuracy}}
     \psfrag{min}[t]{\small{Minutes}}
     \includegraphics[width=0.49 \textwidth]{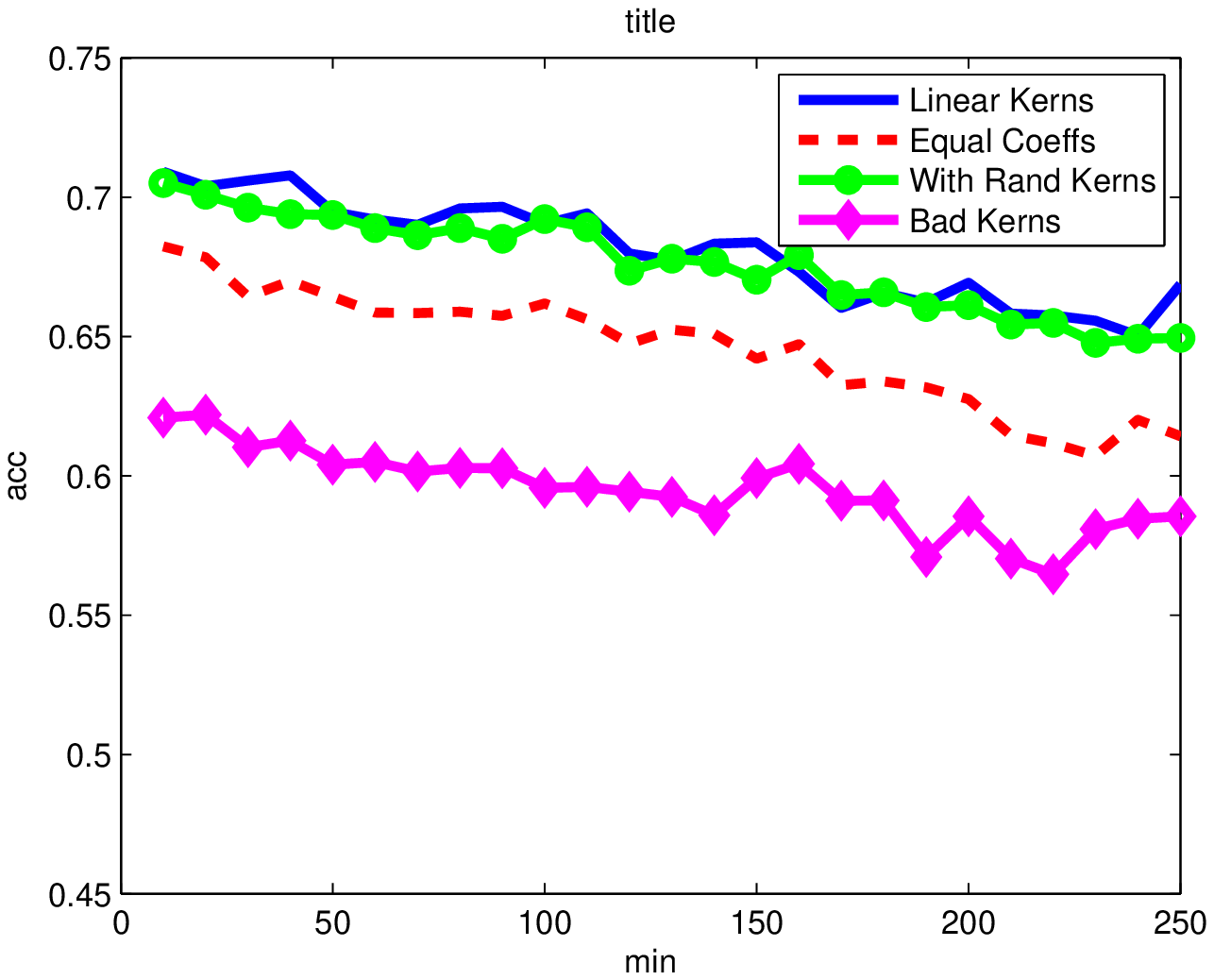}&
    \psfrag{title}[b]{\small{Sharpe Ratio for Various Tests}}
    \psfrag{sharpe}[b]{\small{Sharpe Ratio}}
    \psfrag{min}[t]{\small{Minutes}}
    \includegraphics[width=0.49 \textwidth]{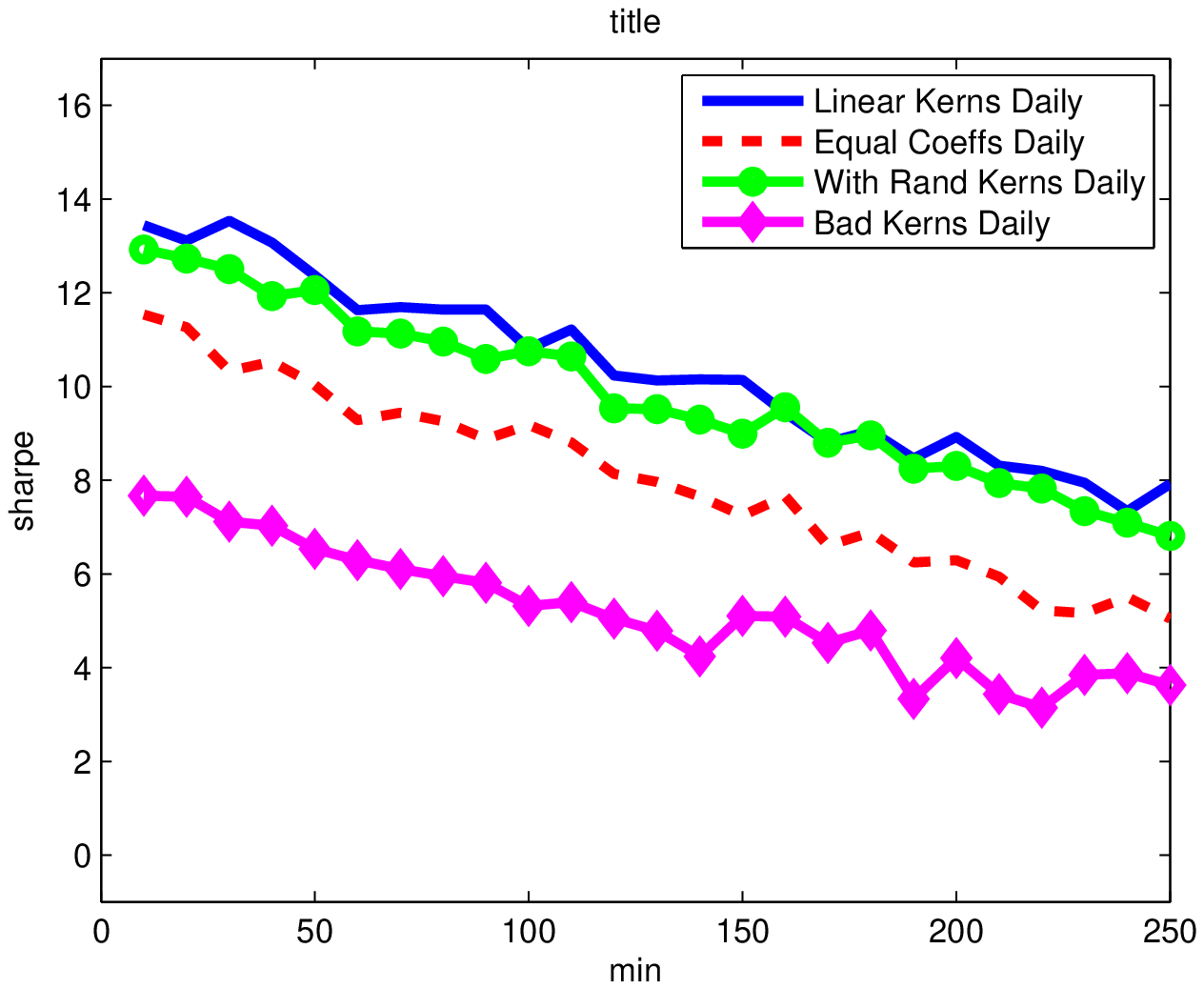}
  \end{tabular}
\caption{Accuracy and Sharpe Ratio for MKL with different kernel sets.  \emph{Linear Kerns} uses 4 linear kernels.  \emph{Equal Coeffs} uses 13 equally weighted kernels.  \emph{With Rand Kerns} adds 3 random kernels to 13 kernels.  \emph{Bad Kerns} uses 4 gaussian kernels with misspecified constants (2 text and 2 absolute returns).  The $75^{th}$ percentile is used as threshold to define abnormal returns.}
\label{fig:various_kernel_tests}
\end{center} \end{figure}

Figure \ref{fig:various_kernel_tests} displays the accuracy and Sharpe ratios of these experiments. Performance using only linear kernels is high since linear kernels achieved equivalent performance to gaussian kernels using SVM.  Adding three random kernels to the mix of thirteen kernels that achieve high performance does not significantly impact the results either.  The three random kernels have negligible coefficients across the horizon (not displayed).  A noticeable decrease in performance is seen when using equally weighted kernels, while an even more significant decrease is observed when using highly suboptimal kernels.  A small data set (using only data after 11 pm) showed an even smaller decrease in performance with equally weighted kernels.  This demonstrates that MKL need not be solved to a high tolerance in order to achieve good performance in this application, while it is still, as expected, necessary to include good kernels in the mix.

\section{Conclusion}
\label{sec:conclusion}

We found significant performance when predicting abnormal returns using text and absolute returns as features.  In addition, multiple kernel learning was introduced to this application and greatly improved performance.  Finally, a cutting plane algorithm for solving large-scale MKL problems was described and its efficiency relative to current MKL solvers was demonstrated.

These experiments could of course be further refined by implementing a tradeable strategy based on abnormal return predictions such as done for daily predictions in Section \ref{subsec:daily_movements}.  Unfortunately, while equity options are liquid assets and would produce realistic performance metrics, intraday options prices are not publicly available.

An important direction for further research is feature selection, i.e. choosing the words in the dictionary.  The above experiments use a simple handpicked set of words.  Techniques such as recursive feature elimination (RFE-SVM) were used to select words but performance was similar to results when using the handpicked dictionary.  More advanced methods such as latent semantic analysis, probabilistic latent semantic analysis, and latent dirichlet allocation should be considered.  Additionally, industry-specific dictionaries can be developed and used with the associated subset of companies.


Another natural extension of our work is regression analysis.  Support vector regressions (SVR) are the regression counterpart to SVM and extend to MKL.  Text can be combined with returns in order to forecast both intraday volatility and abnormal returns using SVR and MKL.

\subsection*{Acknowledgements}
The authors are grateful to Jonathan Lange and Kevin Fan for superb research assistance. We would also like to acknowledge support from NSF grant DMS-0625352, NSF CDI grant SES-0835550, a NSF CAREER award, a Peek junior faculty fellowship and a Howard B. Wentz Jr. junior faculty award.

\small
\bibliographystyle{agsm}
\bibliography{svmFinance}

\end{document}